\pdfoutput=1
\documentclass[11pt]{article}

\usepackage[preprint]{acl}
\usepackage{times}
\usepackage{latexsym}

\usepackage[T1]{fontenc}
\usepackage[utf8]{inputenc}

\usepackage{microtype}

\usepackage{inconsolata}

\usepackage[bulgarian,english]{babel}
\usepackage{graphicx}
\usepackage{tabularx}
\usepackage{placeins}
\usepackage{longtable}
\usepackage[T1]{fontenc}
\usepackage{array}
\usepackage{lipsum} 
\usepackage{subcaption}

\newcolumntype{P}[1]{>{\raggedright\arraybackslash}p{#1}}

\usepackage[symbol]{footmisc}

\newcommand{\insait}{\textsc{INSAIT}\xspace}
\newcommand{\model}{\textsc{BgGPT-Gemma-2}\xspace}
\newcommand{\modelbig}{\textsc{BgGPT-Gemma-2-27B-Instruct}\xspace}
\newcommand{\modelmed}{\textsc{BgGPT-Gemma-2-9B-Instruct}\xspace}

\usepackage{mathtools}
\usepackage{amsfonts}
\usepackage{dsfont}
\usepackage{xspace}
\usepackage{siunitx}
\usepackage{booktabs}       % professional-quality tables
\usepackage[capitalise]{cleveref}
\usepackage{multirow}

\usepackage{algorithm}
\usepackage[noend]{algorithmic}

\newcolumntype{x}[2]{S[table-format=#1.#2,table-auto-round]}

\usepackage{pifont}% http://ctan.org/pkg/pifont

\newcommand{\mbf}[1]{\mathbf{#1}}

\NewDocumentCommand{\phimodel}{O{}}{%
\textsc{Phi}\ifstrempty{#1}{}{-{#1}}\xspace
}

\NewDocumentCommand{\gemma}{O{}}{%
\textsc{Gemma}\ifstrempty{#1}{}{-{#1}}\xspace
}

\NewDocumentCommand{\llama}{O{}O{}}{%
\textsc{Llama}\ifstrempty{#1}{}{-{#1}}\ifstrempty{#2}{}{-{#2}}\xspace
}

\NewDocumentCommand{\llamainstruct}{O{}O{}}{%
\textsc{Llama}\ifstrempty{#1}{}{-{#1}}-\textsc{Instruct}\ifstrempty{#2}{}{-{#2}}\xspace
}

\NewDocumentCommand{\gpt}{O{}}{%
\textsc{GPT}\ifstrempty{#1}{}{-{#1}}\xspace
}

\NewDocumentCommand{\gptturbo}{O{}}{%
\textsc{GPT}\ifstrempty{#1}{}{-{#1}}-\textsc{Turbo}\xspace
}

\NewDocumentCommand{\mistral}{O{}O{}}{%
\textsc{Mistral}\ifstrempty{#1}{}{-{#1}}\ifstrempty{#2}{}{-v0.{#2}}\xspace
}

\NewDocumentCommand{\mistralinstruct}{O{}O{}}{%
\textsc{Mistral-Instruct}\ifstrempty{#1}{}{-{#1}}\ifstrempty{#2}{}{-v0.{#2}}\xspace
}

\usepackage{tikz}

\usetikzlibrary{arrows}
\usetikzlibrary{automata}
\usetikzlibrary{calc}
\usetikzlibrary{backgrounds}
\usetikzlibrary{decorations.markings}
\usetikzlibrary{decorations.pathmorphing}
\usetikzlibrary{decorations.pathreplacing}
\usetikzlibrary{fit}
\usetikzlibrary{patterns}
\usetikzlibrary{positioning}
\usetikzlibrary{shadows}
\usetikzlibrary{shapes}
\usetikzlibrary{shapes.geometric}
\usetikzlibrary{arrows.meta, calc}

\definecolor{mypurple}{HTML}{613F99}
\definecolor{myblue}{HTML}{0071BC}
\definecolor{mygreen}{HTML}{3C8031}

\title{BgGPT 1.0: \\ Extending English-centric LLMs to other languages}

\author{\hspace{-1em}Anton Alexandrov\footnotemark$^{1}$, Veselin Raychev$^{1,2}$, Dimitar I. Dimitrov$^{1,3}$\\ 
        \textbf{Ce Zhang$^{1,4,5}$, Martin Vechev$^{1,3}$, Kristina Toutanova$^{1}$}\\
	\hspace{-3em}$^{1}$ INSAIT, Sofia University ``St. Kliment Ohridski''\hspace{3em} $^{2}$LogicStar.ai\\ \hspace{-3em}$^{3}$ ETH Zurich\hspace{1em}$^{4}$ University of Chicago\hspace{1em}$^{5}$ Together AI\hspace{1em}  \hfil\\
}

\begin{document}
\maketitle

\begin{abstract}
We present \modelbig and \modelmed{}: continually pretrained and fine-tuned versions of Google's Gemma-2 models, specifically optimized for Bulgarian language understanding and generation.
Leveraging Gemma-2's multilingual capabilities and over 100 billion tokens of Bulgarian and English text data, our models demonstrate strong performance in Bulgarian language tasks, setting a new standard for language-specific AI models.
Our approach maintains the robust capabilities of the original Gemma-2 models, ensuring that the English language performance remains intact.
To preserve the base model capabilities, we incorporate continual learning strategies based on recent Branch-and-Merge techniques as well as thorough curation and selection of training data.
We provide detailed insights into our methodology, including the release of model weights with a commercially-friendly license, enabling broader adoption by researchers, companies, and hobbyists.
Further, we establish a comprehensive set of benchmarks based on non-public educational data sources to evaluate models on Bulgarian language tasks as well as safety and chat capabilities.
Our findings demonstrate the effectiveness of fine-tuning state-of-the-art models like Gemma 2 to enhance language-specific AI applications while maintaining cross-lingual capabilities.
\end{abstract}

\section{Introduction}
\footnotetext[0]{$^*$ Correspondence author: anton.alexandrov@insait.ai}

\begin{table*}[tbhp]
    \centering
    \caption{Performance of open large language models on 10 Bulgarian benchmarks, as well as the average scores for the English version of the bencmarks.}
      \label{tab:main}
    \resizebox{\linewidth}{!}{
    \begin{tabular}{lx{2}{2}x{2}{2}|x{2}{1}x{2}{1}x{2}{1}x{2}{1}x{2}{1}x{2}{1}x{2}{1}x{2}{1}x{2}{1}x{2}{1}}
        \toprule
        Model & {English} & {Bulgarian} & \multicolumn{8}{c}{Standard LLM evaluation benchmarks translated to Bulgarian} & \multicolumn{2}{c}{Native Bulgarian benchmarks} \\
              & {(average)} & {(average)} & {Wino*} & {Hellaswag} & {Arc-C\dag} & {Arc-E\dag} & {MMLU} & {MathQA} & {GSM-8K} & {TriviaQA} & {MON} & {EXAMS} \\
        \midrule
        Gemma-2-2b-it & 57.29 & 34.92 & 57.9 & 45.7 & 31.4 & 42 & 41.7 & 28.6 & 25.9 & 9.5 & 34.2 & 32.3\\
        \midrule
        Mistral-7B-Instruct-v0.2 & 61.54 & 43.14 & 60.7 & 56.4 & 39.3 & 55.2 & 47.3 & 32.5 & 29.7 & 24.5 & 43.5 & 42.2\\
        BgGPT-Instruct-7B-v0.2 & 62.89 & 54.36 & 69.7 & 69.6 & 49.7 & 69.5 & 53.8 & 35.2 & 56.9 & 40.9 & 48.9& 49.3\\
        Mistral-Nemo-Instruct-2407 (12B) & 69.11 & 50.24 & 66.8 & 56.7 & 42.2 & 59.6 & 56 & 35.6 & 58.8 & 31.6 &45.0& 50.1\\
        Qwen2.5-7B-Instruct & 62.22 & 40.2 & 57.7 & 47.1 & 34 & 45.3 & 57.6 & 39.7 & 35.3 & 10.4 & 38.4 & 36.4\\
        Llama-3.1-8B-Instruct & 67.89 & 46.69 & 60.6 & 52.8 & 36.9 & 54.9 & 52.5 & 38.4 & 57.6 & 25.5 & 42.2 & 45.5\\
        \midrule
        Gemma-2-9b-it & 70.92 & 55.36 & 70 & 64.6 & 48.2 & 64.6 & 63.1 & 39.2 & 75.7 & 31.3 & 47.3 & 49.6\\
        \textbf{\modelmed} & 71.53 & 61.27 & 72.6 & 71.5 & 54.2 & 75 & 64.4 & 45 & 73.8 & 43.6 & 52.7 & $\mbf{59.7}$\\
        \midrule
        Gemma-2-27b-it & 75.08 & 60.51 & 72.1 & 71 & 49.9 & 63.4 & 69.8 & 43.1 & 81 & 46.8 & 51.2& 56.7\\
        \textbf{\modelbig} & 75.12 & $\mbf{64.67}$ & $\mbf{73.5}$ & $\mbf{74.9}$ & $\mbf{59.3}$ & $\mbf{78.1}$ & 69.8 & 47 & 77.3 & $\mbf{52.9}$ & $\mbf{54.2}$ & 59.2 \\
        \midrule
        Qwen2.5-72B-Instruct & 77.53 & 58.25 & 67.4 & 66.2 & 44.3 & 56.5 & $\mbf{78.8}$ & $\mbf{58.0}$ & $\mbf{81.7}$ & 31.8 & 46.0 & 51.8\\
        Llama-3.1-70B-Instruct & $\mbf{77.75}$ & 61.49 & 67.6 & 67.3 & 52.4 & 70.2 & 75.3 & 53.9 & 68.5 & 51.2 & 52.2 & 56.1\\
        \bottomrule
        \multicolumn{10}{l}{* - Winogrande challenge, \dag - ARC-Challenge and ARC-Easy}
        \end{tabular}
    }
\end{table*}

Large Language Models (LLMs) have shown remarkable capabilities, particularly in the English language, due to the abundance of digitalized English text. Moreover, with recent developments ~\citep{grattafiori2024llama3herdmodels, gemma2, qwen2.5, qwen2} we are seeing that the open-source/weight community has been closing the gap compared to closed, black box platforms in most tasks.
A large portion of LLM users, however, are not native English speakers or require text generation in low-resource languages, and English-centric models do not always meet their needs. Some such models do exhibit strong multi-lingual understanding, stemming from their large, diverse pretraining data, especially in more popular European languages, but often do not meet the required standard for text generation nowadays.

This work presents the development process and evaluation of our language models, specialized for Bulgarian and English use, named \model, based on the Gemma-2 architecture~\citep{gemma2}.
This is the third iteration of Bulgarian models developed by \insait, designed to be powerful yet accessible on common consumer hardware.

\paragraph{Advancing Bulgarian while retaining base capabilities}
The \model models are trained through continued pretraining of strong base models not specialized in Bulgarian.
Our training data also includes English data but has an emphasis on Bulgarian local knowledge and language understanding.
Since the base models demonstrate strong performance in English, coding, mathematics, and other domains, we have applied Branch-and-Merge~\citep{alexandrov-etal-2024-mitigating}, which significantly reduces catastrophic forgetting from the base model while minimally impacting the learning of Bulgarian capabilities.
Our training procedure also incorporates techniques such as curriculum learning and experience replay~\citep{ScialomCM22,ZhangF0N23,ibrahim2024simple}.

\paragraph{Challenges and goals}
Previous releases of open models such as Llama, Llama-2, Mistral 7B and even GPT-2 ~\cite{touvron2023llamaopenefficientfoundation,touvron2023llama, jiang2023mistral7b, Radford2019LanguageMA} have led to numerous specialized models targeting non-English languages~\cite{LeoLM, vikhr, vo2024redwhaleadaptedkoreanllm, turkish, vinallamallamabasedvietnamesefoundation, efficientlyadaptingpretrainedlanguage, OpenLLMRomanian, YugoGPT, fauno}.

Even though the best open-weight foundational models are mostly English-centric, they exhibit reasonable multi-lingual understanding and even often manage to generate, from a linguistic point of view, excellent text in other languages. We find that with the emergence of more capable models like Llama-3.1 and Gemma-2, language-specific fine-tuning has become more demanding in order to produce significant improvements without decreasing the base English performance~\citep{nanda}.
Our primary objective with \model is to enhance the Bulgarian capabilities of a state-of-the-art model while maintaining its strong performance across other tasks.
At the same time, we observe that even powerful models exhibit reduced mathematical, few-shot, knowledge retrieval and other capabilities when prompted in Bulgarian (see section \ref{sec:collapse}), instead of English.
Additionally, sometimes these models default to English responses or exhibit language confusion \citep{languageconfusion} for Bulgarian prompts.
This work focuses on addressing these challenges to produce the best existing open Bulgarian models.

\paragraph{Model capabilities}
Our primary results for Bulgarian are presented in Table \ref{tab:main}. We evaluate on a set of 10 standard benchmarks, translated to Bulgarian using machine translation with manual edits and corrections by professional translators, or coming from authentically Bulgarian sources.
Each benchmark has a maximum score of $100.0$, which would correspond to all tasks being correctly completed.
Eight of the benchmarks contain multiple-choice questions, whose answers are computed using token probability and two evaluate the models' ability to generate text.
Since log probabilities are typically not available for commercial models, we only include open models in this table.
The benchmarks are publicly available \footnote{\url{https://github.com/insait-institute/lm-evaluation-harness-bg}} and can be executed on any open model.
We provide more details about our benchmarking settings in Section \ref{sec:eval}.

Overall, \model performs well on Bulgarian tasks, surpassing even larger models such as Qwen-2.5-72B ~\citep{qwen2.5} and Llama-3.1-70B in specific benchmarks.
Among models larger than \modelbig, we observe varying levels of Bulgarian proficiency.
Specifically, Qwen-2.5 excels in mathematics, coding, and logical reasoning, while Llama-3.1 demonstrates superior performance in general language understanding and knowledge.
At approximately $27$B parameters, \modelbig achieves the highest average performance on Bulgarian tasks while maintaining most of the English capabilities of its base model.
In the $7$-$12$B parameter range, \modelmed outperforms all models on Bulgarian tasks and also overall on English ones.

\paragraph{Applications}
The \model models are available for download with a license permitting personal and commercial use free of charge, including further fine-tuning.
They power the Bulgarian chat service at \url{https://bggpt.ai/}, providing instant access to users without GPU resources.
We anticipate education to be an important application domain for \model. Consequently, we have conducted benchmarking on educational content.
A comparative analysis of these models against state-of-the-art alternatives for chat and educational applications is presented in Section \ref{sec:chat}.

\section{Self-supervised pretraining}

\begin{table*}[tbhp]
    \centering
    \caption{Training process of \modelbig starting from Gemma-2-27b }
      \label{tab:training}
    \resizebox{\linewidth}{!}{
    \begin{tabular}{lx{2}{2}x{2}{2}|x{2}{1}x{2}{1}x{2}{1}x{2}{1}x{2}{1}x{2}{1}x{2}{1}x{2}{1}x{2}{1}x{2}{1}}
        \toprule
        Model & {English} & {Bulgarian} & \multicolumn{8}{c}{Standard LLM evaluation benchmarks translated to Bulgarian} & \multicolumn{2}{c}{Native Bulgarian benchmarks} \\
              & {(average)} & {(average)} & {Wino} & {Hellaswag} & {Arc-C} & {Arc-E} & {MMLU} & {MathQA} & {GSM-8K} & {TriviaQA}  & {MON} & {EXAMS}\\
        \midrule
        Gemma-2-27b & 74.47 & 60.95 & 74 & 72.2 & 52 & 71.7 & 69.3 & 46.9 & 68.7 & 46.3 & 52.2 & 56.2 \\  
        \midrule
        Stage G1 & 74.15 & 62.43 & 74.6 & 73.4 & 51 & 73.1 & 69 & 46 & 71.8 & 52.2 & 52.7 & 60.5 \\
        Stage G2 & 73.36 & 60.57 & 73.5 & 72.5 & 49.5 & 70 & 68.6 & 42.6 & 69.4 & 50.5 & 52.1 & 57 \\
        Merge G1$\oplus$G2 (slerp) & 74.76 & 62.53 & 74.5 & 73.2 & 51.3 & 72.6 & 68.8 & 46 & 72.3 & 52.1 & 53.1 & 61.4 \\
        \midrule
        Stage G3 & 73.99 & 62.43 & 75.1 & 74 & 50.5 & 72.1 & 68.6 & 46.2 & 72.3 & 52.4 & 53.2 & 60 \\
        Stage G4 & 73.25 & 62.01 & 73.9 & 73.6 & 50.7 & 71.5 & 68.9 & 45.1 & 71.7 & 51.8 & 53.4 & 59.5 \\
        Gemma-2-27-it & 75.08 & 60.51 & 72.1 & 71 & 49.9 & 63.4 & 69.8 & 43.1 & 81 & 46.8 & 51.2 & 56.7 \\
        Merge G4$\oplus$IT & 75.99 & 63.97 & 73.2 & 74.4 & 56.1 & 73.1 & 70.6 & 46.4 & 79.7 & 52.3 & 55 & 58.9 \\
        Merge G3$\oplus$(G4$\oplus$IT) & 75.89 & 64.59 & 74.5 & 75.1 & 56.1 & 75.2 & 70.7 & 47.5 & 78.2 & 53.2 & 54.8 & 60.5 \\
        \midrule
        Stage G5 & 73.84 & 63.03 & 77.2 & 74.4 & 51.5 & 72.6 & 69.2 & 45.1 & 73.5 & 53.1 & 53.8 & 60 \\
        Stage G6 & 73.39 & 62.88 & 75.3 & 73.8 & 52.8 & 72.2 & 69.2 & 44.5 & 74.9 & 52.2 & 54 & 60 \\
        G5$\oplus$G6 & 74.04 & 63.42 & 75.7 & 74.2 & 53.1 & 73.1 & 69.8 & 46.4 & 73.9 & 53.4 & 54 & 60.5\\
        \midrule
        Stage G7 & 73.75 & 63.4 & 75.4 & 74.4 & 53 & 73.7 & 69 & 45.9 & 74.8 & 53.2 & 53.9 & 60.8 \\
        Stage G8 & 73.52 & 63.05 & 74.7 & 74.1 & 52.6 & 71.8 & 69.5 & 45.4 & 74.5 & 52.7 & 53.6 & 61.6 \\
        G8$\oplus$IT & 75.87 & 64.35 & 74.3 & 75 & 56.2 & 72.8 & 70.8 & 46.7 & 81.5 & 52.3 & 55.3 & 58.6 \\
        G7$\oplus$(G8$\oplus$IT) = \textbf{base} & 75.46 & 65.06 & 76 & 75.8 & 56.4 & 75.3 & 70.5 & 47.7 & 78.4 & 54.1 & 55.2 & 61.1 \\
        \midrule
        \modelbig= \textbf{base+IFT} & 75.12 & 64.62 & 73.5 & 74.9 & 59.3 & 78.1 & 69.8 & 47 & 77.3 & 52.9 & 54.2 & 59.2 \\
        ((base+BG\_IFT)$\oplus$(base+EN\_IFT))$\oplus$(base+IFT) & 75.93 & 65.49 & 73.56 & 75.42 & 59.64 & 77.82 & 70.27 & 48.27 & 78.32 & 53.81 & 56.5 & 61.4\\
        \bottomrule
        \end{tabular}
    }
\end{table*}

Our models are based on continued pretraining of the Gemma-2 base model with the Branch-and-Merge~\citep{alexandrov-etal-2024-mitigating} strategy.
As this is a technique to mitigate forgetting and erroneous divergence from the base model, we are tracking the scores, particularly for benchmarks where Gemma-2 is already strong -- such as math and reasoning.
We split our pretraining dataset into 8 parts, named G1 to G8 for short.
The odd parts - G1, G3, G5 and G7 contain predominantly Bulgarian text, while the even parts - G2, G4, G6 and G8 contain slightly more English in addition to Bulgarian with the goal of maintaining and possibly improving English skills.
This procedure, described in greater detail in ~\citet{alexandrov-etal-2024-mitigating}, is very similar to what we have already applied for our previous BgGPT models.

In our training, we use PyTorch~\citep{PaszkeGMLBCKLGA19} and HuggingFace's transformers library~\citep{WolfDSCD19} combined with DeepSpeed ZeRO to parallelize our training runs
\citep{deepspeed10.1145/3394486.3406703,zero9355301}.
All training is performed on 64 NVIDIA H100s divided into 8 nodes of 8 GPUs with Infiniband interconnection.
We set $10^{-5}$ as the maximum learning rate for the continued pretraining of all the models combined with a batch size of $512$ for continued pretraining and $256$ or $512$ for supervised fine-tuning.
We use cosine decay to $0.1 \cdot \text{max\_lr}$ with $\max(100,0.01\cdot\text{total\_steps})$ linear warmup. 

\subsection{Pretraining data}
The training data for continued pretaining is completely identical to the one utilized in ~\citet{alexandrov-etal-2024-mitigating}, combining approximate experience replay and high-quality Bulgarian data from various sources. To adapt the RedPajama-V2 ~\cite{weber2024redpajama} annotation pipeline we performed a few adjustments due to the use of the Cyrillic alphabet and the lack of a Bulgarian equivalent of certain resources.
Any natural-language signal that looks at characters has to be adjusted to include Cyrillic and some additional punctuation like a specific type of quotation marks appearing in digitalized Bulgarian text. The original pipeline includes a count per document of words that are contained in the LDNOOBW blocklist, which we replace with our own list of "bad words" in Bulgarian.
RedPajama-V2 also considers ML-based heuristics from fastText ~\citep{fasttext} classifiers trained to distinguish between unfiltered web text and high-quality domains. For English, those are Wikipedia, Books and OpenWebText, from which only Wikipedia is available in Bulgarian. To enrich our set for Bulgarian we replace the English books dataset with Bulgarian books and OpenWebText with a mixture of our high-quality datasets (excluding Wikipedia), such as the parliament proceedings, news articles, legal text and others.
From the three perplexity splits we only consider the "head" and "middle" ones, which contain a combined total of more than 100M documents classified as Bulgarian web page content. To filter on the quality data we compute the statistics of each signal and through manual inspection decide the cut-off percentiles for each, where usually the acceptance ranges between the 5th and 95th percentile for most metrics, unless we consider this to be "too soft". After filtering and deduplication we are left with 46M documents of high-quality Bulgarian web text.

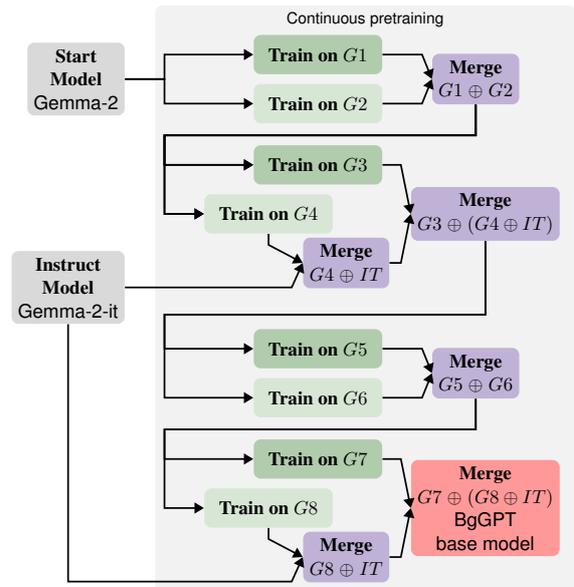
\begin{figure}[t]
\centering
    
\scalebox{0.65}{
\begin{tikzpicture}[node distance=1cm and 1cm, every node/.style={font=\sffamily}]

    \def\dy{1.3cm}
    \def\ddx{0.125cm}
    \def\ddy{0.25cm}

    \def\dwidth{1.1cm}
    \def\dheight{0.8cm}
    \def\boxwidth{1.6cm}

    \node (main) [rounded corners, thick, minimum width=8.5cm, minimum height=12cm, fill=gray!10] at (0,0) {};
    % Title
    \node at ($(main.north) + (0,-0.3cm)$) {\small Continuous pretraining};

    \node (base) [rounded corners, minimum height=\dy, minimum width=\boxwidth, fill=gray!30, anchor=east, align=center] at ($(main.west) + (-0.6, 4.5)$) {\bf Start \\\bf Model \\ Gemma-2};
    \node (it) [rounded corners, minimum height=\dy, minimum width=\boxwidth, fill=gray!30, anchor=east, align=center] at ($(main.west) + (-0.6, 0.25)$) {\bf Instruct \\\bf Model \\ Gemma-2-it};    

    \node (GOne) [rounded corners, minimum height=\dheight, minimum width=2.6cm, fill=mygreen!40, anchor=west, align=center] at ($(main.west |- base.east) + (2.0, 0.5)$) {\bf Train on $G1$};
    \node (GTwo) [rounded corners, minimum height=\dheight, minimum width=2.6cm, fill=mygreen!20, anchor=west, align=center] at ($(GOne.west |- base.east) + (0.0, -0.5)$) {\bf Train on $G2$};

    \draw[-{Triangle},draw=black, line width=1pt] (base.east) -- ($(main.west |- base.east) + (0.2cm,0.0cm)$) -- ($(main.west |- GOne.west) + (0.2cm,0.0cm)$) -- (GOne.west);
    \draw[-{Triangle},draw=black, line width=1pt] (base.east) -- ($(main.west |- base.east) + (0.2cm,0.0cm)$) -- ($(main.west |- GTwo.west) + (0.2cm,0.0cm)$) -- (GTwo.west);

    \node (MergeOT) [rounded corners, minimum height=\dheight, minimum width=1.6cm, fill=mypurple!40, anchor=west, align=center] at ($(GOne.east)!0.5!(GTwo.east) + (1.0,0.0)$) {\bf Merge \\ $G1$ $\oplus$ $G2$};
    \draw[-{Triangle},draw=black, line width=1pt] (GOne.east) -- ($(MergeOT.west |- GOne.east) + (-0.2cm,0.0cm)$) -- (MergeOT.west);
    \draw[-{Triangle},draw=black, line width=1pt] (GTwo.east) -- ($(MergeOT.west |- GTwo.east) + (-0.2cm,0.0cm)$) -- (MergeOT.west);

    \node (GThree) [rounded corners, minimum height=\dheight, minimum width=2.6cm, fill=mygreen!40, anchor=west, align=center] at ($(GOne.west |- base.east) + (0.0, -1.75)$) {\bf Train on $G3$};
    \node (GFour) [rounded corners, minimum height=\dheight, minimum width=2.6cm, fill=mygreen!20, anchor=west, align=center] at ($(main.west |- base.east) + (1.0, -2.75)$) {\bf Train on $G4$};

    \draw[-{Triangle},draw=black, line width=1pt] (MergeOT.south) -- ($(MergeOT.south) + (0,-0.625cm)$) -- ($(main.west |- MergeOT.south) + (0.2cm,-0.625cm)$) -- ($(main.west |- GThree.west) + (0.2cm,0.0cm)$) -- (GThree.west);
    \draw[-{Triangle},draw=black, line width=1pt] (MergeOT.south) -- ($(MergeOT.south) + (0,-0.625cm)$) -- ($(main.west |- MergeOT.south) + (0.2cm,-0.625cm)$) -- ($(main.west |- GFour.west) + (0.2cm,0.0cm)$) -- (GFour.west);    

    \node (MergeTF) [rounded corners, minimum height=\dheight, minimum width=1.6cm, fill=mypurple!40, anchor=west, align=center] at ($(GOne.west |- base.east) + (1.0, -3.75)$) {\bf Merge \\ $G4$ $\oplus$ $IT$};

    % A line from it to MergeTF.
    \draw[-{Triangle},draw=black, line width=1pt] (it.east) -- ($(MergeTF.west |- it.east) + (-0.225cm,0.0cm)$) -- (MergeTF.west);
    % A line from GFour south to MergeTF.
    \draw[-{Triangle},draw=black, line width=1pt] (GFour.south) -- ($(GFour.south) + (0,-0.225cm)$) -- (MergeTF.west);

    \node (MergeTFTF) [rounded corners, minimum height=\dheight, minimum width=1.6cm, fill=mypurple!40, anchor=west, align=center] at ($(GThree.east)!0.5!(MergeTF.east) + (0.5,0.0)$) {\bf Merge \\ $G3$ $\oplus$ $(G4 \oplus IT)$};
    \draw[-{Triangle},draw=black, line width=1pt] (GThree.east) -- ($(MergeTFTF.west |- GThree.east) + (-0.2cm,0.0cm)$) -- (MergeTFTF.west);
    \draw[-{Triangle},draw=black, line width=1pt] (MergeTF.east) -- ($(MergeTFTF.west |- MergeTF.east) + (-0.2cm,0.0cm)$) -- (MergeTFTF.west);

    \node (GFive) [rounded corners, minimum height=\dheight, minimum width=2.6cm, fill=mygreen!40, anchor=west, align=center] at ($(GOne.west |- base.east) + (0.0, -5.5)$) {\bf Train on $G5$};
    \node (GSix) [rounded corners, minimum height=\dheight, minimum width=2.6cm, fill=mygreen!20, anchor=west, align=center] at ($(GOne.west |- base.east) + (0.0, -6.5)$) {\bf Train on $G6$};

    \draw[-{Triangle},draw=black, line width=1pt] (MergeTFTF.south) -- ($(MergeTFTF.south) + (0,-1.5cm)$) -- ($(main.west |- MergeTFTF.south) + (0.2cm,-1.5cm)$) -- ($(main.west |- GFive.west) + (0.2cm,0.0cm)$) -- (GFive.west);
    \draw[-{Triangle},draw=black, line width=1pt] (MergeTFTF.south) -- ($(MergeTFTF.south) + (0,-1.5cm)$) -- ($(main.west |- MergeTFTF.south) + (0.2cm,-1.5cm)$) -- ($(main.west |- GSix.west) + (0.2cm,0.0cm)$) -- (GSix.west);
    
    \node (MergeFS) [rounded corners, minimum height=\dheight, minimum width=1.6cm, fill=mypurple!40, anchor=west, align=center] at ($(GFive.east)!0.5!(GSix.east) + (1.0,0.0)$) {\bf Merge \\ $G5$ $\oplus$ $G6$};
    \draw[-{Triangle},draw=black, line width=1pt] (GFive.east) -- ($(MergeFS.west |- GFive.east) + (-0.2cm,0.0cm)$) -- (MergeFS.west);
    \draw[-{Triangle},draw=black, line width=1pt] (GSix.east) -- ($(MergeFS.west |- GSix.east) + (-0.2cm,0.0cm)$) -- (MergeFS.west);

    \node (GSeven) [rounded corners, minimum height=\dheight, minimum width=2.6cm, fill=mygreen!40, anchor=west, align=center] at ($(GOne.west |- base.east) + (0.0, -7.75)$) {\bf Train on $G7$};
    \node (GEight) [rounded corners, minimum height=\dheight, minimum width=2.6cm, fill=mygreen!20, anchor=west, align=center] at ($(main.west |- base.east) + (1.0, -8.75)$) {\bf Train on $G8$};

    \draw[-{Triangle},draw=black, line width=1pt] (MergeFS.south) -- ($(MergeFS.south) + (0,-0.625cm)$) -- ($(main.west |- MergeFS.south) + (0.2cm,-0.625cm)$) -- ($(main.west |- GSeven.west) + (0.2cm,0.0cm)$) -- (GSeven.west);
    \draw[-{Triangle},draw=black, line width=1pt] (MergeFS.south) -- ($(MergeFS.south) + (0,-0.625cm)$) -- ($(main.west |- MergeFS.south) + (0.2cm,-0.625cm)$) -- ($(main.west |- GEight.west) + (0.2cm,0.0cm)$) -- (GEight.west);

    \draw[-{Triangle},draw=black, line width=1pt] (MergeOT.south) -- ($(MergeOT.south) + (0,-0.625cm)$) -- ($(main.west |- MergeOT.south) + (0.2cm,-0.625cm)$) -- ($(main.west |- GThree.west) + (0.2cm,0.0cm)$) -- (GThree.west);
    \draw[-{Triangle},draw=black, line width=1pt] (MergeOT.south) -- ($(MergeOT.south) + (0,-0.625cm)$) -- ($(main.west |- MergeOT.south) + (0.2cm,-0.625cm)$) -- ($(main.west |- GFour.west) + (0.2cm,0.0cm)$) -- (GFour.west);

    \node (MergeSE) [rounded corners, minimum height=\dheight, minimum width=1.6cm, fill=mypurple!40, anchor=west, align=center] at ($(GOne.west |- base.east) + (1.0, -9.75)$) {\bf Merge \\ $G8$ $\oplus$ $IT$};

    % A line from it to MergeSE.
    \draw[-{Triangle},draw=black, line width=1pt] (it.south) -- ($(it.south |- MergeSE.west) + (0.0cm,-0.5cm)$) -- ($(MergeSE.west) + (-0.225cm,-0.5cm)$) -- (MergeSE.west);
    % A line from GEight south to MergeSE.
    \draw[-{Triangle},draw=black, line width=1pt] (GEight.south) -- ($(GEight.south) + (0,-0.225cm)$) -- (MergeSE.west);

    \node (MergeSESE) [rounded corners, minimum height=\dheight, minimum width=1.6cm, fill=red!40, anchor=west, align=center] at ($(GSeven.east)!0.5!(MergeSE.east) + (0.5,0.0)$) {\bf Merge \\ $G7$ $\oplus$ $(G8 \oplus IT)$ \\ BgGPT \\ base model};
    \draw[-{Triangle},draw=black, line width=1pt] (GSeven.east) -- ($(MergeSESE.west |- GSeven.east) + (-0.2cm,0.0cm)$) -- (MergeSESE.west);
    \draw[-{Triangle},draw=black, line width=1pt] (MergeSE.east) -- ($(MergeSESE.west |- MergeSE.east) + (-0.2cm,0.0cm)$) -- (MergeSESE.west);

\end{tikzpicture}
}
\caption{Overview of the \model training process. The training data is split into 8 parts denoted G1 to G8 and the process involves training on data (green boxes) and merging (denoted with $\oplus$).}
\label{fig:overview}
\end{figure}

\subsection{Training procedure}
We summarize the training steps in Figure \ref{fig:overview}, alongside their evaluation performance in Table \ref{tab:training}.
We begin by continually pretraining the base Gemma-2-27b model on the G1 and G2 datasets separately.
Training on both datasets results in an overall increase in Bulgarian skills and a reduction in English skills.
By merging the G1 and G2 models, we obtain a model denoted G1$\oplus$G2 that recovers its English skills.
We use the SLERP procedure from Arcee's MergeKit~\citep{goddard2024arcee} to merge the models, as we have found it to perform best compared to other merging methods in terms of the learning-forgetting trade-off. Other methods may be better at regularization but at the cost of a reduced compound improvement on the target domain.
We search for ratios between $0.3$ and $0.7$ to find the best merge among the two models. The best merges are chosen based on the average benchmark performance, where we prioritize Bulgarian scores but also try to maximize English scores. For example, if the $0.5$ merge has a 0.1\% higher score in Bulgarian but 0.5\% lower score in English compared to the $0.6$ merge, then we would choose the latter.

\paragraph{Continued pretraining}
Starting from G1$\oplus$G2, we train the next two models on G3 and G4.
At this stage, instead of merging only G3 and G4, we first merge G4 with Gemma-2-27b-it, as G4 was trained on more English data, and then this merged model is merged again with G3.
The goal of deviation from BaM is to inject skills stemming from the instruction tuning stage of Gemma-2-27b-it. We observe that Gemma-2-27b-it is very capable in Bulgarian, with little language confusion. Since the instruction fine-tuning dataset for that model is not publicly available, we find that merging it with our models provides a possibility to insert those capabilities into our model without massively disrupting the training process.

As visible in Table \ref{tab:training}, the merge operation with Gemma-2-27b-it brings significant math skills and helps us avoid loss in others.
The resulting model denoted G3$\oplus$(G4$\oplus$IT) contains a quarter of the effective weight from Gemma-2-27b-it.

The next steps with G5 and G6 are performed in the same manner as G1 and G2, but this time starting from G3$\oplus$(G4$\oplus$IT). Finally, we continually train G5$\oplus$G6 on the G7 and G8 datasets.
We repeat the merge strategy from earlier, with the G7 model being merged with the merge of G8 and Gemma-2-27b-it.

Overall, this training procedure results in a base model that performs well across all tasks and brings skills from the instruction-tuned version of Gemma-2.
For additional details on the trajectories of English benchmarks, see Table~\ref{tab:forgetting}.

\section{Supervised fine-tuning}

\subsection{Instruction fine-tuning data}
For instruction fine-tuning, we collect a set of translated and native Bulgarian conversations and instructions that are summarized in Table \ref{tab:bg_ift_data}.

\paragraph{Translated datasets}
Some datasets were translated from English using the Google Translate API. We also manually fixed translations that we detected as likely incorrect using heuristics. Our heuristics search for inconsistencies in the English and Bulgarian translations as the alphabets are different, e.g. if a word is kept in English in one sentence and translated or transliterated in the next one, we flag the example for manual translation.
Overall, as the cost of translation is high, the amount of Bulgarian instruction data is more limited than native English data.

\paragraph{Chat conversations}
We also leverage the chat application at \url{https://bggpt.ai/} and collect a set of conversations generated by current users.
We gather conversations from consenting users of our chat application to construct an IFT dataset. As part of this data creation process, we removed sensitive user information. We analyze the diversity of the prompts from users and assign a category/topic to each one. To do this, we first manually go through 2000 random prompts and assign them a category of our choice, we then organize the chaotic labeling into 82 different topics that we consider to be common or worth adding. After that, we prompt a language model to label the queries one by one by choosing one of the given categories. In Figure ~\ref{fig:chat_topics_ift} we show a distribution of all 82 topics assigned as relevant to the queries from the conversations and in Figure~\ref{fig:chat_supercat_ift} we can see how they are distributed into supercategories. As we can see most prompts are related to science and education and content generation, while there is also a decent chunk of questions associated with Bulgaria (history, literature, geography, people, etc).

\paragraph{Toxic questions}

We assemble a set of unsafe, toxic, and inappropriate questions to create a denial QA dataset. In particular, we first translate the filtered chat logs using INSAIT's previous model, BgGPT-Instruct-v0.2, into English. This allows us to use existing open-weight LLMs, to identify toxic and inappropriate questions regardless of the language (mostly Bulgarian and English). We employ Mistral-7B-v0.2 for identification, as empirically it worked well to detect toxic content due to its alignment training. We then apply exact and near deduplication of the resulting unsafe queries based on embedding similarity. We give this deduplicated set of queries to our annotators to divide into different topics, further filter for good-quality examples, and finally supply denial answers to them. This gives us a total of 550 question-answer pairs to include in our IFT dataset. 

\paragraph{Human preference dataset}
In our endeavors to apply Direct Preference Optimization ~\citep{rafailov2023direct} for Human Alignment, before the launch of the first chat application, we let more than $70$ people create all kinds of complex queries and gathered a total of 1500 prompts, which we then generated answers for with BgGPT-Instruct-v0.2, a few different versions of Bulgarian models we trained and other models. Our annotators then expressed their preference for one of the possible generations if there was an appropriate one or edited the one that was closest to being acceptable and set that as the preferred answer. As this process is time-consuming and the improvement gained from DPO on such a small amount of data was insignificant, we did not continue in this direction. We found that simply including the question and preferred answer as IFT pairs was helpful enough to improve the style and instruction-following capabilities of the model.

\paragraph{Synthetic poetry questions}
Rhyming and poetry are very difficult skills for a language model to obtain because they combine creative writing with generation constraints that are based on subwords and potential pronunciations, which is something tokenizer-based LMs notoriously have a hard time with. We found that our first models of the BgGPT series (BgGPT-Instruct-v0.2), which were fine-tuned mostly on translated instructions, really struggled to make any rhymes at all, often actually trying to rhyme words that would rhyme in English but when translated into Bulgarian would not. To tackle this we used a rhyme dictionary and annotated thousands of songs songs and poems. We then only took rhyming line cuts or stanzas to create questions and answers such as "Complete the given stanza and retain the rhyming form \{stanza lines $(0,k-j)$\} => \{stanza lines $(k-j,k)$\}", "Give $\{k\}$ words that rhyme with the word \{word\} => \{extracted rhyming words\}", "Which of the following word pairs rhyme?" and others. We found that including too many of these examples in the IFT stage can lead to overfitting but adding even a small amount helps the model improve in creating rhyming songs and poems. 
\begin{table}[t]
    \centering
    \caption{Composition of the Bulgarian IFT dataset.}
    \label{tab:bg_ift_data}
    \resizebox{\linewidth}{!}{
    \begin{tabular}{llc}
        \toprule
        Dataset & Domain & \#Examples  \\
        \midrule
        \multicolumn{3}{l}{English datasets} \\
        ~~SlimOrca + MetaMath & Mixed Conversations & $100,000$ \\
        \midrule
        \multicolumn{3}{l}{Datasets translated from English} \\
        ~~Capybara-BG & Mixed Conversations & $16,000$ \\
        ~~MetaMath-BG & Math & $10,000$ \\
        ~~CodeAlpaca-BG & Code & $5,000$ \\
        \midrule
        \multicolumn{3}{l}{Datasets obtained from \url{https://bggpt.ai/}} or synthetically \\
        ~~BgDPO & Preference & $1500$ \\
        ~~BgGPT Toxic Data & Toxicity & $500$ \\
        ~~Chat log questions & Mixed Conversations & $140,000$ \\
        \multicolumn{2}{l}{~~synthetic rhymes, songs, poems} & $1,000$ \\
        \bottomrule
    \end{tabular}
    }
\end{table}

\subsection{Instruction fine-tuning process}
For our final chat-oriented model we simply fine-tune for 2 epochs on the dataset composition described in \ref{tab:bg_ift_data}. However, this is not the best-performing model in terms of benchmark scores, but the one we find to match our preference, as well as the model with the highest preference of GPT4o-as-a-judge. Safety, answering style and helpfulness were preferred in this case, as this is the model that we incorporate into our public chat system.
Nevertheless, in our Branch-and-Merge studies~\cite{alexandrov-etal-2024-mitigating}, we find that BaM is very effective in the IFT stage when one splits and merges based on the language of the instructions. This is also visible in the last row of Table~\ref{tab:training}, where we employ a more complex scheme, compared to the straightforward tuning. We fine-tune one model on ~800K combined conversations from SlimOrca, MetaMath and Capybara, which becomes the English branch. Another is trained on all of the available translated datasets - Capybara-BG (16K), SlimOrca-BG (50K), MetaMath-BG (50K), CodeAlpaca-BG (5k), which becomes the Bulgarian branch, that is afterward merged with the English one. This merged model is then merged with the aforementioned best chat model. This final merge becomes the best-performing model on almost all benchmarks.

\section{Benchmark Evaluation}\label{sec:eval}

For our evaluations, we used and now elaborate on the benchmarks applied in LM Evaluation Harness~\citep{eval-harness}.
We translated them to Bulgarian using Google Translate API and then applied heuristics to identify samples that may be translated incorrectly.
Then, we asked professional translators to review and correct these translations.
Additionally, we utilize 2 benchmarks that were originally created in Bulgarian, namely EXAMS~\citep{exams} and a novel benchmark created from official exam questions from the Bulgarian Ministry of Education (described more thoroughly in Section ~\ref{sec:chat}).

\paragraph{Winogrande challenge} \citep{winogrande} is a commonsense reasoning benchmark measuring an LLM's ability to fill in a blank from a choice of two entities to logically complete a sentence. The set of tasks in reference are the 1767 samples from the validation set of \texttt{winogrande\_xl}. As part of our translation effort, some samples were rephrased as the word-to-word Bulgarian version could reveal the answer through the gender of the words from the rest of the sentence.
We evaluate the models on this benchmark in a 0-shot manner.

\paragraph{HellaSwag} \citep{hellaswag} is a commonsense reasoning benchmark evaluating how well an LLM can select a logical continuation of a sentence. It is run on the 10000-sample validation set using 0-shot prompting.

\paragraph{ARC-Easy and -Challenge} \citep{arc} is a dataset of science exam questions. We evaluate on the 2590 hard samples (ARC-Challenge) and 5197 easy samples (ARC-Easy) in a 0-shot manner.

\paragraph{MMLU} \citep{mmlu} is a multitask language understanding benchmark covering a wide range of 57 different tasks. Evaluated on 14079 test set samples in a 5-shot manner.

\paragraph{MathQA} \citep{mathqa} is a multiple choice mathematical reasoning benchmark. Evaluated on the 4475 validation set samples.

\paragraph{GSM8K} \citep{gsm8k} is a mathematical reasoning benchmark consisting of grade-school math questions for which free-text answers must be provided. Evaluated on 1.3k test set samples. We run GSM8k with a 5-shot chain-of-thought generation.

\paragraph{TriviaQA} \citep{triviaqa} is a dataset consisting of trivia questions. We evaluate 17.9k validation set samples in a 5-shot manner.

\paragraph{EXAMS} \citep{exams} is a high school exam question dataset covering a range of subjects. Evaluated on 1472 test set samples in Bulgarian. We use 5-shot prompting.

\paragraph{Additional Benchmarks}
In our evaluation, we additionally used the following benchmarks but omitted them from this report. While being of relatively high quality, we found that the Belebele~\citep{belebele} benchmark was very simple, with little to no difference in the performance between models, except the distinction between the scores of IFT-d models and non-IFT-d ones. We also made use of XNLI~\citep{xnli} for Bulgarian, as provided in the LM Evaluation Harness, however, we found that the examples in Bulgarian were very confusing and unnaturally worded. The quality may have also been reflected in the scores we observed, which seemed very random, with no sign of what drove the numbers higher or lower. While models that have seen significant Bulgarian training performed better than their English-centric bases, we found that all of them scored around 50\%, unrelated to how much or what type of Bulgarian training they have received.

\section{Instruction-following Evaluation}\label{sec:chat}

\begin{table*}[tbhp]
    \centering
    \caption{Performance on school test exams provided by the Ministry of Education and Science of Bulgaria, turned into open questions, judged by gpt4o for correctness. The correct and incorrect answers are given to the judge.}
      \label{tab:monjudge}
    \resizebox{\linewidth}{!}{
    \begin{tabular}{lx{2}{2}|x{2}{2}x{2}{2}x{2}{2}x{2}{2}x{2}{2}x{2}{2}x{2}{2}x{2}{2}x{2}{2}}
        \toprule
              &               & \multicolumn{7}{c}{Per subject breakdown of grades} \\
        Model & {Average}  & {Literature} & {Math} & {Geography \&} & {History} & {IT skills} & {Physics \&} & {Human \&} \\
              &  {grade (up to 6)}    &              &        & {Economics}    &           &             & {Astronomy}  & {Nature}   \\

        \midrule
        \modelbig                   &	5.41  &  4.93 & 5.53 & 5.22 & 5.42 & 5.6 & 5.66 & 5.62 \\
        Gemma-2-27B-it           &   5.28  &  4.55 & 5.47 & 5.02 & 5.12 & 5.56 & 5.67 & 5.54 \\
        \midrule
        GPT-4o-mini-2024-07-18          &	5.38  &  4.78 & 5.59 & 5.11 & 5.24 & 5.67 & 5.74 & 5.61 \\
        GPT-4o-2024-08-06               &	5.58  &  5.22 & 5.68 & 5.39 & 5.64 & 5.72 & 5.85 & 5.7 \\
        \midrule
        Llama-3.1-405B-Instruct-Turbo (*)  &	5.36  &  4.61 & 5.59 & 5.21 & 5.48 & 5.59 & 5.71 & 5.52 \\
        Llama-3.1-70B-Instruct-Turbo (*)   &	5.2  &  4.43 & 5.43 & 5.04 & 5.25 & 5.48 & 5.57 & 5.43 \\
        Llama-3.3-70B-Instruct-Turbo (*)   &    5.18  &  4.41 & 5.38 & 5.02 & 5.23 & 5.43 & 5.5 & 5.39 \\
        Qwen2.5-72B-Instruct-Turbo (*)     &	5.29  &  4.46 & 5.69 & 4.97 & 5.19 & 5.7 & 5.76 & 5.54 \\
        \bottomrule
        (*) - Served by a chat API endpoint by TogetherAI.
        \end{tabular}
    }
\end{table*}

In this section, we evaluate the utility of the model in chat between a user and an AI assistant users are expected to interact with via its API endpoint or chat application.
For its chat interactions, \model follows the same chat template as the instruction-tuned Gemma-2 models.

Many of the English and Multilingual benchmarks in such a setting use human judges~\citep{lmarena} or model judges~\citep{ChatbotArena2023}.
However, we find it challenging and costly to simply translate these benchmarks to a new language like Bulgarian and maintain their quality.
Instead, we propose a set of new benchmarks based on local Bulgarian data that we collected from a variety of sources.

\subsection{MON benchmark}\label{sec:collapse}

MON is a dataset collected by the Ministry of Education and Science of Bulgaria~\citeyear{mon}, which contains a set of multiple choice questions for school exams between 4th and 12th grade.
These exams are designed as a standardized way to assess the knowledge of students across Bulgaria on a range of topics.
These questions are in Bulgarian and cover subjects like literature, mathematics, physics, history, computer science, and all other subjects in the standard school curriculum in Bulgaria.
We include only text-based questions, without images or other media, with exactly one correct answer out of 4 and as a result, obtain $10,088$ questions with 4 possible answers.
The tests were carefully designed and taken by thousands of students; we thus believe that the dataset is of relatively high quality.
At this stage, we do not intend to release the dataset publicly, to avoid potential future contamination of models.

To evaluate the quality of the models in a chat setting with educational questions, we first converted the multiple-choice questions into open-text questions using GPT-4o. We noticed that GPT-4o failed to transform some of the examples that were questions specifically referring to the options, which were now omitted, or required additional text or other form of media that was missing. We constructed a prompt with elaborate instructions and 20 examples with reasoning for GPT-4o to identify such questions, which it easily did and then removed those. 
With the resulting $8,828$ questions, we prompted all models to provide answers and then used GPT-4o to judge their correctness.
Based on each model's response and the known correct answer, the judge assigned a grade according to the Bulgarian grading system, where 2 is the lowest and 6 is the highest score. It is important to note that some of the converted questions, from closed to open, become harder to judge the correctness of, as the correct choice from the given 4 options may become one of many possible right answers, or simply become a part of the expected output, with the question requiring a more verbose and thorough answer. With this in mind we do not exclude the possibility of bias in the judge's scoring, however, we believe it is reduced to a minimum.  We show some examples of the pipelines in Table ~\ref{tab:mon_examples}.

We evaluate on this benchmark using the models' chat API endpoints, allowing us to compare to proprietary models.
Overall, we find that \modelbig performs above the level of GPT-4o-mini, outperforming it in literature and knowledge-related tasks while under-performing on reasoning tasks.
In comparison to Llama-3.1, \modelbig outperforms on average even the biggest Llama-3.1-405B model, but not in all subjects as it is a much smaller model. A notable other model is Qwen-2.5-72B, which is particularly strong in math, but is weaker in other areas. We note that both Llama-3.1 and Qwen-2.5 are generally powerful models, but they often respond in incorrect Bulgarian or with language confusion (which may not always be picked up by the judge), which makes them undesirable for Bulgarian text generation The results are shown in Table~\ref{tab:monjudge}.

\subsection{Chat Preference with a Judge Model}

\begin{figure}[t]
    \centering
    \includegraphics[width=0.52\textwidth]{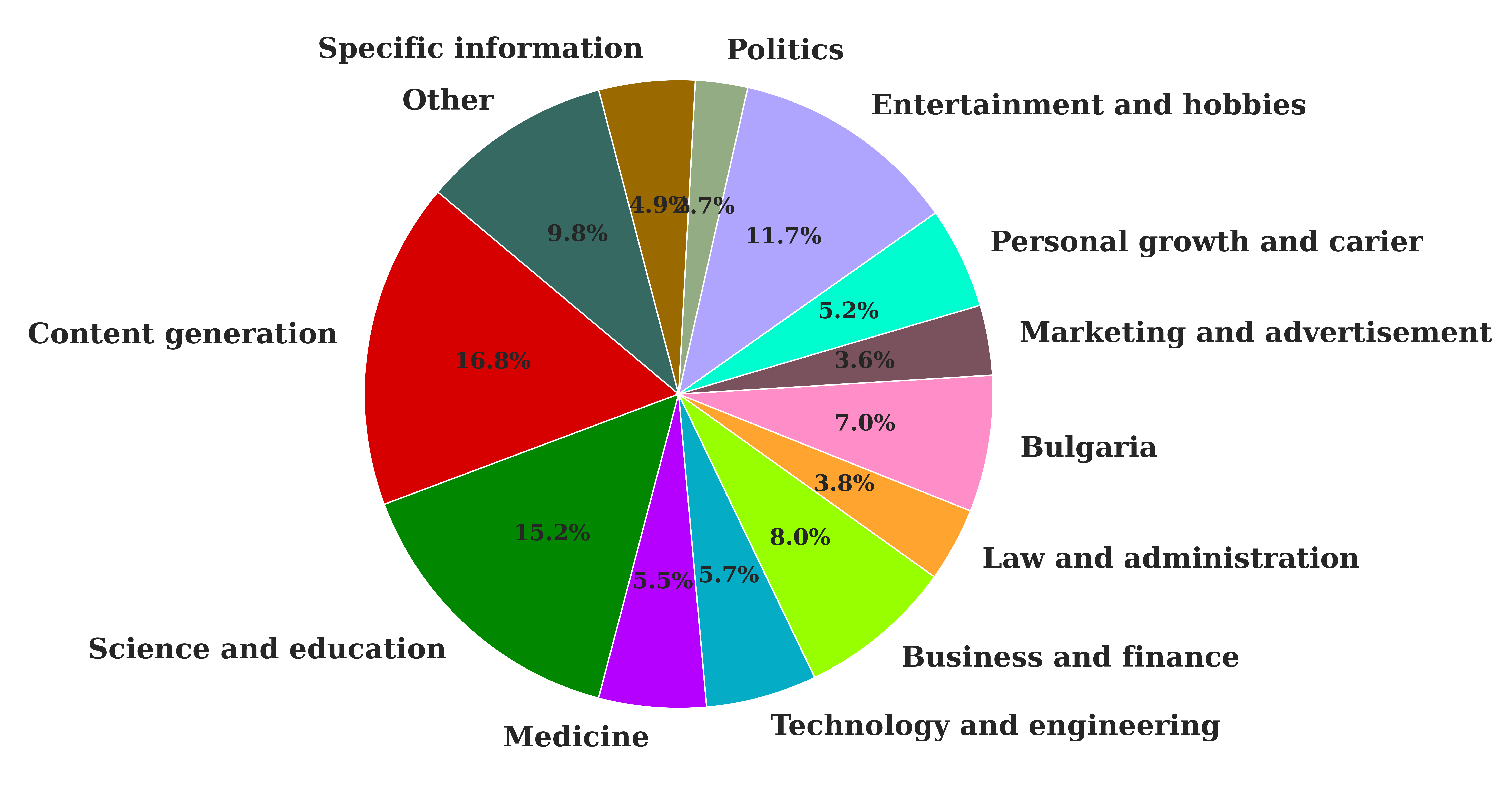}
    \caption{Distribution of the supercategories of the assigned main topics of the preference benchmark questions.}
    \label{fig:chat_supercat_eval_inpaper}
\end{figure}

We have found that even though GPT4o may not be perfect for Bulgarian text generation, it can serve as a great LLM-as-a-judge to give preference between two outputs from user prompts. Taking advantage of the already existing bggpt.ai chat application we collect a dataset of chat interactions from our consenting users to create a set of 3000 prompts to use the preference judge on. We apply various quality filters to exclude erroneous examples like single-word prompts and other nonsensical input. We deduplicate the examples based on embedding similarity and take action to remove inputs including sensitive or personally identifiable information. As we aim to evaluate our models on a diverse set of representative user queries, we classify the examples into 82 different topic categories and soft sample our 3000 queries to include all topics while maintaining the diversity distribution.  
It is important to note that these are sampled out of a separate chunk of conversations and not from those used within the IFT dataset. We can see a distribution of the supercategories of these prompts in Figure ~\ref{fig:chat_supercat_eval_inpaper} and a fine-grained 82-topic distribution in Figure ~\ref{fig:chat_topics_eval}.

\begin{figure*}[h]
    \centering
    \includegraphics[width=0.9\linewidth]{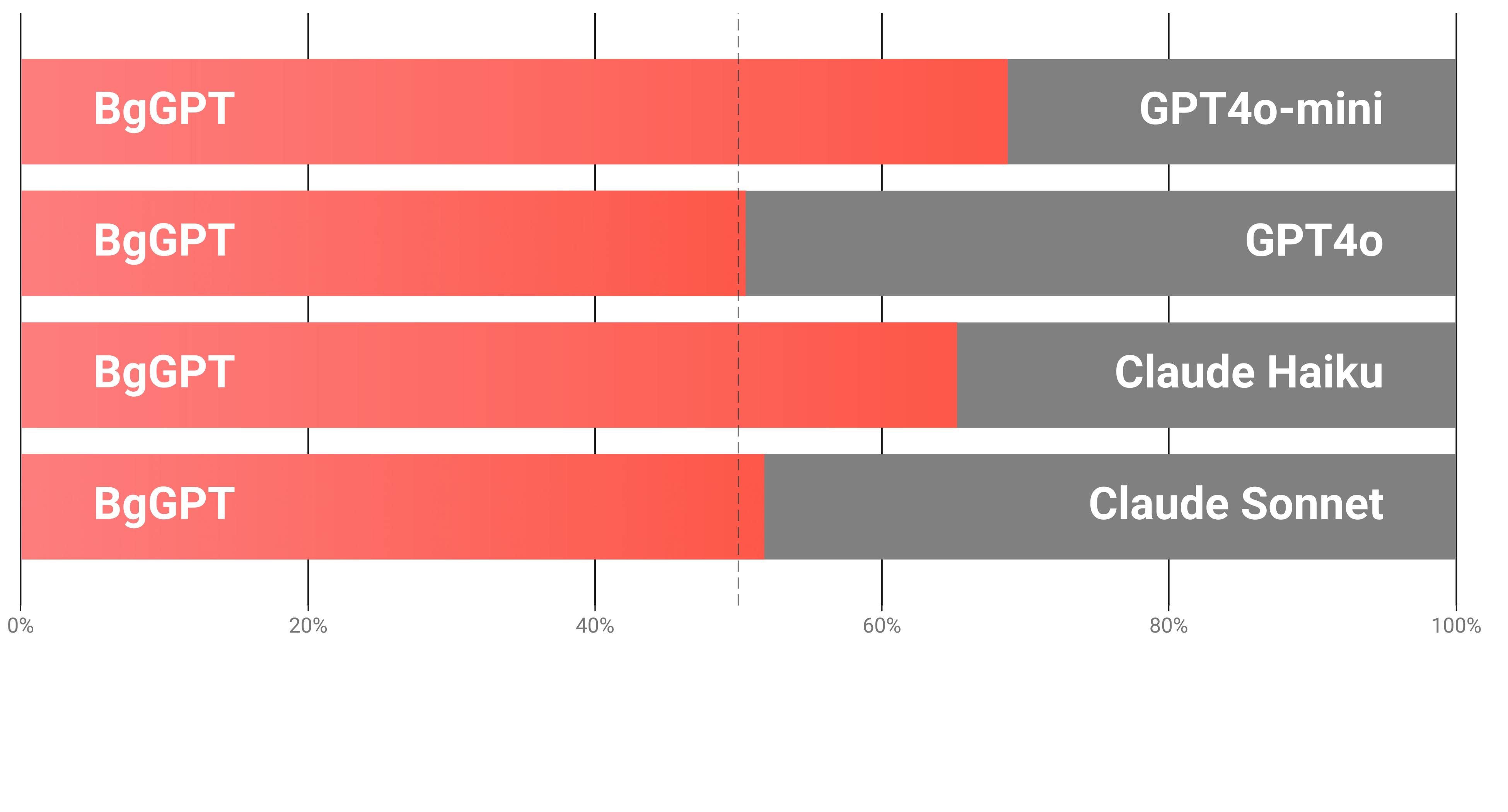}
    \vspace{-4.5em}
    \caption{Preference balance on diverse real user queries of \modelbig against SOTA closed chat models from OpenAI and Anthropic.}
    \label{fig:preference}
\end{figure*}

Experimenting with multiple prompts for the judge in both Bulgarian and English, we found that GPT4o's better abilities to reason in English resulted in better judgments when supplying the judge instructions in English. Further, we found that the Wildbench~\citep{lin2024wildbench} judge prompt reduces the positional bias of the judge for Bulgarian compared to the alternatives, while still producing good judgements, thus we chose it for our experiments. To reduce the positional bias further, we allow for only 3 possible judgements - a preference for the first answer, a preference for the second answer or a tie, and ran all judgements with both possible orderings of the models' outputs. We awarded 0 pts for a tie and 0.5 pts for a preference for a particular model per direction, allowing for maximum of 1 pts per question to be awarded for each model. We show the results in \cref{fig:preference}, where we find that GPT-4o judges our model's answers to be significantly more preferable than GPT-4o-mini's answers and on par or slightly more preferable than its own answers. 

\section{Related work}
Our major goal is to create language models that are capable of understanding and generating both Bulgarian and English, which is what we believe would be most useful to propagate the advances of NLP in Bulgarian locality.
The techniques used to create strong duolingual language models can be divided into 2 distinct types - ones that construct a model from scratch and those that build on top of existing language models. The focus of our work is on the latter approach, leveraging the already invested resources of the increasingly competitive open-source and open-weight LLM research.
It has been shown that transfer learning from one domain to another can not only reduce the costs of model development but also bring improved performance on the target domain ~\citep{zhuang2020comprehensivesurveytransferlearning, anonymous2024scaling}. More specifically, in the domain of language modeling, there is evidence of cross-lingual relevance, as models can utilize their capabilities through different languages, particularly for low-resource languages~\citep{conneau-etal-2020-unsupervised, wang-etal-2024-probing-emergence, he2024scalinglawsmultilinguallanguage}. However, this is not without caveats, such as the \emph{curse of multilinguality} ~\citep{conneau-etal-2020-unsupervised} and \emph{catastrophic forgetting} ~\citep{french1999catastrophic,GoodfellowMDCB13,GogoulouLBN23,ShiXWQWWW24}. To overcome these challenges we only work with 2 languages and employ different strategies to alleviate catastrophic forgetting in language transfer. 

\paragraph{Language transfer}
There is an abundance of works that show very promising results of language transfer from an English-centric base model to a target language of lower resource ~\citep{LeoLM, vinallamallamabasedvietnamesefoundation, efficientlyadaptingpretrainedlanguage, nanda, turkish, joshi2024adaptingmultilingualllmslowresource, vo2024redwhaleadaptedkoreanllm, luukkonen-etal-2023-fingpt} through different strategies. Most examples focus on improving on relatively weaker baselines, such as Llama-2~\citep{touvron2023llama}, or vastly struggle with retaining English capabilities. Recent model releases such as the Gemma-2 model family ~\citep{gemma2} demonstrate very strong multilingual performance even on lower resource languages such as Bulgarian, which sets a higher new standard for bilingual model improvement.

Some works rely on extending or altering the architecture of the underlying transformer model through added model weights, adapters, swapped layers or vocabulary extension~\citep{nanda, gurgurov-etal-2024-adapting, anonymous2024layer, vinallamallamabasedvietnamesefoundation}, which is something we avoid due to the increased complexity of deployment and use for the community.
It is made obvious that targeting a single language with a monolingual corpus reduces the original English proficiency and that experience/knowledge replay can alleviate it to a great extent~\citep{llamaenglishempiricalstudy, ibrahim2024simple, efficientlyadaptingpretrainedlanguage}, however, due to the unavailable pretraining data we use approximate knowledge replay.

\section{Conclusion}

We presented our entire \model pipeline for training strong duolingual language models, illustrating the steps for the Bulgarian language. The resulting pipeline has been used to build the best open Bulgarian language model to date.

Our findings emphasize how to leverage strong English-centric open-weight models through domain-specific adaptation strategies. Specifically, our \modelbig model exhibits state-of-the-art performance on Bulgarian benchmarks, outperforming larger, multilingually capable models like Qwen-2.5-72B and Llama-3.1-70B on crucial metrics.

We highlight the challenges involved in achieving high-quality generation in the target language of interest (e.g., Bulgarian) while maintaining English proficiency and the already learned capabilities.
This work provides detailed insights into the necessary data construction and curation required to train such models, as well as the continual pretraining and fine-tuning strategies, including leveraging the latest research on model merging (i.e., our Branch-and-Merge method), that effectively preserve and enhance the bilingual capabilities of the models. 

We demonstrate the performance of the \model models on educational tasks, real-world chat applications in addition to a multitude of standard benchmarks, translated and naturally generated. While exhibiting strong performance in all of our Bulgarian evaluations, they preserve or even improve upon the already competitive English benchmark scores.
This achievement demonstrates the viability of constructing robust, specialized LLMs that address new linguistic domains.

The insights gained from this study lay the groundwork for future advancements in the low-resource language modeling and multilingual representation learning areas. Applications such as Bulgarian educational tools, chat interfaces, and domain-specific knowledge retrieval systems stand to benefit significantly. The public accessibility of \model ensures that these enhancements can be extended beyond academic research, promoting broader adoption and practical use, coming from the high performance-to-size ratio.
In our ongoing research we aim to close the gap between the performance of open-weight models on tasks in various languages and provide state-of-the-art methods for language specific NLP.

% \clearpage

\section{Limitations}
While we show that our methods are efficient in creating duolingual models, we rely on the existing multilingual capabilities of the base models. Extending open-weight models like Gemma-2 and Llama-3 to German and Bulgarian is reasonable but such success cannot be guaranteed for significantly lower-resource languages and may require further cross-lingual advancements to compensate for the lack of training data.

In our fine-tuning efforts we find that there is a lack of high-quality data and while we have put a great effort in filling that gap, it is far from enough and more work is needed to improve the various aspects of this training stage.

This work describes the development of large language model, which is one of the components for building chat assistants, smart applications and tools that may be applied in scenarios of various criticality. It is important to note that generative language models based on decoder transformers may not be reliable and should be thoroughly tested and safety bounded subject to the specific use case.

\section{Ethical Considerations}
We believe our work empowers the general public that uses Bulgarian to benefit from the power of large language models. However, such models can of course also be abused and in particular if our approach generalizes beyond language to general domain adaptation, malicious practitioners could more efficiently adapt the models for nefarious tasks.

\section*{Acknowledgments}
This research was partially funded by the Ministry of Education and Science of Bulgaria (support for INSAIT, part of the Bulgarian National Roadmap for Research Infrastructure).

This work has been done as part of the EU grant ELSA (European Lighthouse on Secure and Safe AI, grant agreement no. 101070617). Views and opinions expressed are however those of the authors only and do not necessarily reflect those of the European Union or European Commission. Neither the European Union nor the European Commission can be held responsible for them.

The work has received funding from the Swiss State Secretariat for Education, Research and Innovation (SERI).

We would like to thank Dr. Maurice Weber for helping adapt the RPv2 pipeline to Bulgarian and create the Bulgarian dataset. We would also like to thank Hristo Venev for his help with system-related issues.

% \input{latex/sections/style_instructions}

\clearpage

\bibliography{references}

\begin{thebibliography}{60}
\providecommand{\natexlab}[1]{#1}

\bibitem[{Aleksa(2024)}]{YugoGPT}
Gordi{\'c} Aleksa. 2024.
\newblock \href {https://huggingface.co/gordicaleksa/YugoGPT} {Yugogpt - an open-source llm for serbian, bosnian, and croatian languages}.

\bibitem[{Alexandrov et~al.(2024)Alexandrov, Raychev, Mueller, Zhang, Vechev, and Toutanova}]{alexandrov-etal-2024-mitigating}
Anton Alexandrov, Veselin Raychev, Mark~Niklas Mueller, Ce~Zhang, Martin Vechev, and Kristina Toutanova. 2024.
\newblock \href {https://aclanthology.org/2024.findings-emnlp.1000} {Mitigating catastrophic forgetting in language transfer via model merging}.
\newblock In \emph{Findings of the Association for Computational Linguistics: EMNLP 2024}, pages 17167--17186, Miami, Florida, USA. Association for Computational Linguistics.

\bibitem[{Amini et~al.(2019)Amini, Gabriel, Lin, Koncel-Kedziorski, Choi, and Hajishirzi}]{mathqa}
Aida Amini, Saadia Gabriel, Shanchuan Lin, Rik Koncel-Kedziorski, Yejin Choi, and Hannaneh Hajishirzi. 2019.
\newblock \href {https://doi.org/10.18653/v1/N19-1245} {{M}ath{QA}: Towards interpretable math word problem solving with operation-based formalisms}.
\newblock In \emph{Proceedings of the 2019 Conference of the North {A}merican Chapter of the Association for Computational Linguistics: Human Language Technologies, Volume 1 (Long and Short Papers)}, pages 2357--2367, Minneapolis, Minnesota. Association for Computational Linguistics.

\bibitem[{Anonymous(2024{\natexlab{a}})}]{anonymous2024layer}
Anonymous. 2024{\natexlab{a}}.
\newblock \href {https://openreview.net/forum?id=vQhn4wrQ6j} {Layer swapping for zero-shot cross-lingual transfer in large language models}.
\newblock In \emph{Submitted to The Thirteenth International Conference on Learning Representations}.
\newblock Under review.

\bibitem[{Anonymous(2024{\natexlab{b}})}]{anonymous2024scaling}
Anonymous. 2024{\natexlab{b}}.
\newblock \href {https://openreview.net/forum?id=vPOMTkmSiu} {Scaling laws for downstream task performance in machine translation}.
\newblock In \emph{Submitted to The Thirteenth International Conference on Learning Representations}.
\newblock Under review.

\bibitem[{Arena(2024)}]{lmarena}
Chatbot Arena. 2024.
\newblock \href {https://lmarena.ai/} {Chatbot arena (formerly lmsys)}.

\bibitem[{Bacciu et~al.(2023)Bacciu, Trappolini, Santilli, Rodolà, and Silvestri}]{fauno}
Andrea Bacciu, Giovanni Trappolini, Andrea Santilli, Emanuele Rodolà, and Fabrizio Silvestri. 2023.
\newblock \href {https://arxiv.org/abs/2306.14457} {Fauno: The italian large language model that will leave you senza parole!}
\newblock \emph{Preprint}, arXiv:2306.14457.

\bibitem[{Bandarkar et~al.(2023)Bandarkar, Liang, Muller, Artetxe, Shukla, Husa, Goyal, Krishnan, Zettlemoyer, and Khabsa}]{belebele}
Lucas Bandarkar, Davis Liang, Benjamin Muller, Mikel Artetxe, Satya~Narayan Shukla, Donald Husa, Naman Goyal, Abhinandan Krishnan, Luke Zettlemoyer, and Madian Khabsa. 2023.
\newblock \href {https://arxiv.org/abs/2308.16884} {The belebele benchmark: a parallel reading comprehension dataset in 122 language variants}.
\newblock \emph{Preprint}, arXiv:2308.16884.

\bibitem[{Bojanowski et~al.(2017)Bojanowski, Grave, Joulin, and Mikolov}]{fasttext}
Piotr Bojanowski, Edouard Grave, Armand Joulin, and Tomas Mikolov. 2017.
\newblock \href {https://arxiv.org/abs/1607.04606} {Enriching word vectors with subword information}.
\newblock \emph{Preprint}, arXiv:1607.04606.

\bibitem[{Choudhury et~al.(2024)Choudhury, Chauhan1, Das, Sahnan, Han, Li, Singh, Jadhav, Agarwal, Choudhary, Banerjee, Koto, Bhat, Shukla, Ghosh, Kamboj, Pandit, Pradhan, Mullah, Pal, Sahu, Doraiswamy, Sengupta, Ramakrishnan, Joshi, Gosal, Sheinin, Vassilieva, and Nakov}]{nanda}
Monojit Choudhury, Shivam Chauhan1, Rocktim~Jyoti Das, Dhruv Sahnan, Xudong Han, Haonan Li, Aaryamonvikram Singh, Alok~Anil Jadhav, Utkarsh Agarwal, Mukund Choudhary, Debopriyo Banerjee, Fajri Koto, Junaid Bhat, Awantika Shukla, Samujjwal Ghosh, Samta Kamboj, Onkar Pandit, Lalit Pradhan, Parvez Mullah, Rahul Pal, Sunil Sahu, Soundar Doraiswamy, Neha Sengupta, Gokul Ramakrishnan, Rituraj Joshi, Gurpreet Gosal, Avraham Sheinin, Natalia Vassilieva, and Preslav Nakov. 2024.
\newblock \href {https://github.com/mbzuai-nlp/Llama-3-Nanda-10B-Chat} {Llama-3-nanda-10b-chat: An open generative large language model for hindi}.

\bibitem[{Clark et~al.(2018)Clark, Cowhey, Etzioni, Khot, Sabharwal, Schoenick, and Tafjord}]{arc}
Peter Clark, Isaac Cowhey, Oren Etzioni, Tushar Khot, Ashish Sabharwal, Carissa Schoenick, and Oyvind Tafjord. 2018.
\newblock \href {https://api.semanticscholar.org/CorpusID:3922816} {Think you have solved question answering? try arc, the ai2 reasoning challenge}.
\newblock \emph{ArXiv}, abs/1803.05457.

\bibitem[{Cobbe et~al.(2021)Cobbe, Kosaraju, Bavarian, Chen, Jun, Kaiser, Plappert, Tworek, Hilton, Nakano, Hesse, and Schulman}]{gsm8k}
Karl Cobbe, Vineet Kosaraju, Mohammad Bavarian, Mark Chen, Heewoo Jun, Lukasz Kaiser, Matthias Plappert, Jerry Tworek, Jacob Hilton, Reiichiro Nakano, Christopher Hesse, and John Schulman. 2021.
\newblock \href {https://api.semanticscholar.org/CorpusID:239998651} {Training verifiers to solve math word problems}.
\newblock \emph{ArXiv}, abs/2110.14168.

\bibitem[{Conneau et~al.(2020)Conneau, Khandelwal, Goyal, Chaudhary, Wenzek, Guzm{\'a}n, Grave, Ott, Zettlemoyer, and Stoyanov}]{conneau-etal-2020-unsupervised}
Alexis Conneau, Kartikay Khandelwal, Naman Goyal, Vishrav Chaudhary, Guillaume Wenzek, Francisco Guzm{\'a}n, Edouard Grave, Myle Ott, Luke Zettlemoyer, and Veselin Stoyanov. 2020.
\newblock \href {https://doi.org/10.18653/v1/2020.acl-main.747} {Unsupervised cross-lingual representation learning at scale}.
\newblock In \emph{Proceedings of the 58th Annual Meeting of the Association for Computational Linguistics}, pages 8440--8451, Online. Association for Computational Linguistics.

\bibitem[{Conneau et~al.(2018)Conneau, Rinott, Lample, Williams, Bowman, Schwenk, and Stoyanov}]{xnli}
Alexis Conneau, Ruty Rinott, Guillaume Lample, Adina Williams, Samuel Bowman, Holger Schwenk, and Veselin Stoyanov. 2018.
\newblock \href {https://doi.org/10.18653/v1/D18-1269} {{XNLI}: Evaluating cross-lingual sentence representations}.
\newblock In \emph{Proceedings of the 2018 Conference on Empirical Methods in Natural Language Processing}, pages 2475--2485, Brussels, Belgium. Association for Computational Linguistics.

\bibitem[{Csaki et~al.(2023)Csaki, Pawakapan, Thakker, and Xu}]{efficientlyadaptingpretrainedlanguage}
Zoltan Csaki, Pian Pawakapan, Urmish Thakker, and Qiantong Xu. 2023.
\newblock \href {https://arxiv.org/abs/2311.05741} {Efficiently adapting pretrained language models to new languages}.
\newblock \emph{Preprint}, arXiv:2311.05741.

\bibitem[{French(1999)}]{french1999catastrophic}
Robert~M French. 1999.
\newblock Catastrophic forgetting in connectionist networks.
\newblock \emph{Trends in cognitive sciences}, 3(4):128--135.

\bibitem[{Gao et~al.(2023)Gao, Tow, Abbasi, Biderman, Black, DiPofi, Foster, Golding, Hsu, Le~Noac'h, Li, McDonell, Muennighoff, Ociepa, Phang, Reynolds, Schoelkopf, Skowron, Sutawika, Tang, Thite, Wang, Wang, and Zou}]{eval-harness}
Leo Gao, Jonathan Tow, Baber Abbasi, Stella Biderman, Sid Black, Anthony DiPofi, Charles Foster, Laurence Golding, Jeffrey Hsu, Alain Le~Noac'h, Haonan Li, Kyle McDonell, Niklas Muennighoff, Chris Ociepa, Jason Phang, Laria Reynolds, Hailey Schoelkopf, Aviya Skowron, Lintang Sutawika, Eric Tang, Anish Thite, Ben Wang, Kevin Wang, and Andy Zou. 2023.
\newblock \href {https://doi.org/10.5281/zenodo.10256836} {A framework for few-shot language model evaluation}.

\bibitem[{Gemma-Team et~al.(2024)Gemma-Team, Riviere, Pathak, Sessa, Hardin, Bhupatiraju, Hussenot, Mesnard, Shahriari, Ramé, Ferret, Liu, Tafti, Friesen, Casbon, Ramos, Kumar, Lan, Jerome, Tsitsulin, Vieillard, Stanczyk, Girgin, Momchev, Hoffman, Thakoor, Grill, Neyshabur, Bachem, Walton, Severyn, Parrish, Ahmad, Hutchison, Abdagic, Carl, Shen, Brock, Coenen, Laforge, Paterson, Bastian, Piot, Wu, Royal, Chen, Kumar, Perry, Welty, Choquette-Choo, Sinopalnikov, Weinberger, Vijaykumar, Rogozińska, Herbison, Bandy, Wang, Noland, Moreira, Senter, Eltyshev, Visin, Rasskin, Wei, Cameron, Martins, Hashemi, Klimczak-Plucińska, Batra, Dhand, Nardini, Mein, Zhou, Svensson, Stanway, Chan, Zhou, Carrasqueira, Iljazi, Becker, Fernandez, van Amersfoort, Gordon, Lipschultz, Newlan, yeong Ji, Mohamed, Badola, Black, Millican, McDonell, Nguyen, Sodhia, Greene, Sjoesund, Usui, Sifre, Heuermann, Lago, McNealus, Soares, Kilpatrick, Dixon, Martins, Reid, Singh, Iverson, Görner, Velloso, Wirth, Davidow, Miller, Rahtz, Watson,
  Risdal, Kazemi, Moynihan, Zhang, Kahng, Park, Rahman, Khatwani, Dao, Bardoliwalla, Devanathan, Dumai, Chauhan, Wahltinez, Botarda, Barnes, Barham, Michel, Jin, Georgiev, Culliton, Kuppala, Comanescu, Merhej, Jana, Rokni, Agarwal, Mullins, Saadat, Carthy, Cogan, Perrin, Arnold, Krause, Dai, Garg, Sheth, Ronstrom, Chan, Jordan, Yu, Eccles, Hennigan, Kocisky, Doshi, Jain, Yadav, Meshram, Dharmadhikari, Barkley, Wei, Ye, Han, Kwon, Xu, Shen, Gong, Wei, Cotruta, Kirk, Rao, Giang, Peran, Warkentin, Collins, Barral, Ghahramani, Hadsell, Sculley, Banks, Dragan, Petrov, Vinyals, Dean, Hassabis, Kavukcuoglu, Farabet, Buchatskaya, Borgeaud, Fiedel, Joulin, Kenealy, Dadashi, and Andreev}]{gemma2}
Gemma-Team, Morgane Riviere, Shreya Pathak, Pier~Giuseppe Sessa, Cassidy Hardin, Surya Bhupatiraju, Léonard Hussenot, Thomas Mesnard, Bobak Shahriari, Alexandre Ramé, Johan Ferret, Peter Liu, Pouya Tafti, Abe Friesen, Michelle Casbon, Sabela Ramos, Ravin Kumar, Charline~Le Lan, Sammy Jerome, Anton Tsitsulin, Nino Vieillard, Piotr Stanczyk, Sertan Girgin, Nikola Momchev, Matt Hoffman, Shantanu Thakoor, Jean-Bastien Grill, Behnam Neyshabur, Olivier Bachem, Alanna Walton, Aliaksei Severyn, Alicia Parrish, Aliya Ahmad, Allen Hutchison, Alvin Abdagic, Amanda Carl, Amy Shen, Andy Brock, Andy Coenen, Anthony Laforge, Antonia Paterson, Ben Bastian, Bilal Piot, Bo~Wu, Brandon Royal, Charlie Chen, Chintu Kumar, Chris Perry, Chris Welty, Christopher~A. Choquette-Choo, Danila Sinopalnikov, David Weinberger, Dimple Vijaykumar, Dominika Rogozińska, Dustin Herbison, Elisa Bandy, Emma Wang, Eric Noland, Erica Moreira, Evan Senter, Evgenii Eltyshev, Francesco Visin, Gabriel Rasskin, Gary Wei, Glenn Cameron, Gus Martins,
  Hadi Hashemi, Hanna Klimczak-Plucińska, Harleen Batra, Harsh Dhand, Ivan Nardini, Jacinda Mein, Jack Zhou, James Svensson, Jeff Stanway, Jetha Chan, Jin~Peng Zhou, Joana Carrasqueira, Joana Iljazi, Jocelyn Becker, Joe Fernandez, Joost van Amersfoort, Josh Gordon, Josh Lipschultz, Josh Newlan, Ju~yeong Ji, Kareem Mohamed, Kartikeya Badola, Kat Black, Katie Millican, Keelin McDonell, Kelvin Nguyen, Kiranbir Sodhia, Kish Greene, Lars~Lowe Sjoesund, Lauren Usui, Laurent Sifre, Lena Heuermann, Leticia Lago, Lilly McNealus, Livio~Baldini Soares, Logan Kilpatrick, Lucas Dixon, Luciano Martins, Machel Reid, Manvinder Singh, Mark Iverson, Martin Görner, Mat Velloso, Mateo Wirth, Matt Davidow, Matt Miller, Matthew Rahtz, Matthew Watson, Meg Risdal, Mehran Kazemi, Michael Moynihan, Ming Zhang, Minsuk Kahng, Minwoo Park, Mofi Rahman, Mohit Khatwani, Natalie Dao, Nenshad Bardoliwalla, Nesh Devanathan, Neta Dumai, Nilay Chauhan, Oscar Wahltinez, Pankil Botarda, Parker Barnes, Paul Barham, Paul Michel, Pengchong Jin,
  Petko Georgiev, Phil Culliton, Pradeep Kuppala, Ramona Comanescu, Ramona Merhej, Reena Jana, Reza~Ardeshir Rokni, Rishabh Agarwal, Ryan Mullins, Samaneh Saadat, Sara~Mc Carthy, Sarah Cogan, Sarah Perrin, Sébastien M.~R. Arnold, Sebastian Krause, Shengyang Dai, Shruti Garg, Shruti Sheth, Sue Ronstrom, Susan Chan, Timothy Jordan, Ting Yu, Tom Eccles, Tom Hennigan, Tomas Kocisky, Tulsee Doshi, Vihan Jain, Vikas Yadav, Vilobh Meshram, Vishal Dharmadhikari, Warren Barkley, Wei Wei, Wenming Ye, Woohyun Han, Woosuk Kwon, Xiang Xu, Zhe Shen, Zhitao Gong, Zichuan Wei, Victor Cotruta, Phoebe Kirk, Anand Rao, Minh Giang, Ludovic Peran, Tris Warkentin, Eli Collins, Joelle Barral, Zoubin Ghahramani, Raia Hadsell, D.~Sculley, Jeanine Banks, Anca Dragan, Slav Petrov, Oriol Vinyals, Jeff Dean, Demis Hassabis, Koray Kavukcuoglu, Clement Farabet, Elena Buchatskaya, Sebastian Borgeaud, Noah Fiedel, Armand Joulin, Kathleen Kenealy, Robert Dadashi, and Alek Andreev. 2024.
\newblock \href {https://arxiv.org/abs/2408.00118} {Gemma 2: Improving open language models at a practical size}.
\newblock \emph{Preprint}, arXiv:2408.00118.

\bibitem[{Goddard et~al.(2024)Goddard, Siriwardhana, Ehghaghi, Meyers, Karpukhin, Benedict, McQuade, and Solawetz}]{goddard2024arcee}
Charles Goddard, Shamane Siriwardhana, Malikeh Ehghaghi, Luke Meyers, Vlad Karpukhin, Brian Benedict, Mark McQuade, and Jacob Solawetz. 2024.
\newblock Arcee's mergekit: A toolkit for merging large language models.
\newblock \emph{arXiv preprint arXiv:2403.13257}.

\bibitem[{Gogoulou et~al.(2023)Gogoulou, Lesort, Boman, and Nivre}]{GogoulouLBN23}
Evangelia Gogoulou, Timoth{\'{e}}e Lesort, Magnus Boman, and Joakim Nivre. 2023.
\newblock \href {https://doi.org/10.48550/ARXIV.2311.01200} {A study of continual learning under language shift}.
\newblock \emph{CoRR}, abs/2311.01200.

\bibitem[{Goodfellow et~al.(2014)Goodfellow, Mirza, Da, Courville, and Bengio}]{GoodfellowMDCB13}
Ian~J. Goodfellow, Mehdi Mirza, Xia Da, Aaron~C. Courville, and Yoshua Bengio. 2014.
\newblock \href {http://arxiv.org/abs/1312.6211} {An empirical investigation of catastrophic forgeting in gradient-based neural networks}.
\newblock In \emph{2nd International Conference on Learning Representations, {ICLR} 2014, Banff, AB, Canada, April 14-16, 2014, Conference Track Proceedings}.

\bibitem[{Grattafiori et~al.(2024)Grattafiori, Dubey, Jauhri, Pandey, Kadian, Al-Dahle, Letman, Mathur, Schelten, Vaughan, Yang, Fan, Goyal, Hartshorn, Yang, Mitra, Sravankumar, Korenev, Hinsvark, Rao, Zhang, Rodriguez, Gregerson, Spataru, Roziere, Biron, Tang, Chern, Caucheteux, Nayak, Bi, Marra, McConnell, Keller, Touret, Wu, Wong, Ferrer, Nikolaidis, Allonsius, Song, Pintz, Livshits, Wyatt, Esiobu, Choudhary, Mahajan, Garcia-Olano, Perino, Hupkes, Lakomkin, AlBadawy, Lobanova, Dinan, Smith, Radenovic, Guzmán, Zhang, Synnaeve, Lee, Anderson, Thattai, Nail, Mialon, Pang, Cucurell, Nguyen, Korevaar, Xu, Touvron, Zarov, Ibarra, Kloumann, Misra, Evtimov, Zhang, Copet, Lee, Geffert, Vranes, Park, Mahadeokar, Shah, van~der Linde, Billock, Hong, Lee, Fu, Chi, Huang, Liu, Wang, Yu, Bitton, Spisak, Park, Rocca, Johnstun, Saxe, Jia, Alwala, Prasad, Upasani, Plawiak, Li, Heafield, Stone, El-Arini, Iyer, Malik, Chiu, Bhalla, Lakhotia, Rantala-Yeary, van~der Maaten, Chen, Tan, Jenkins, Martin, Madaan, Malo, Blecher,
  Landzaat, de~Oliveira, Muzzi, Pasupuleti, Singh, Paluri, Kardas, Tsimpoukelli, Oldham, Rita, Pavlova, Kambadur, Lewis, Si, Singh, Hassan, Goyal, Torabi, Bashlykov, Bogoychev, Chatterji, Zhang, Duchenne, Çelebi, Alrassy, Zhang, Li, Vasic, Weng, Bhargava, Dubal, Krishnan, Koura, Xu, He, Dong, Srinivasan, Ganapathy, Calderer, Cabral, Stojnic, Raileanu, Maheswari, Girdhar, Patel, Sauvestre, Polidoro, Sumbaly, Taylor, Silva, Hou, Wang, Hosseini, Chennabasappa, Singh, Bell, Kim, Edunov, Nie, Narang, Raparthy, Shen, Wan, Bhosale, Zhang, Vandenhende, Batra, Whitman, Sootla, Collot, Gururangan, Borodinsky, Herman, Fowler, Sheasha, Georgiou, Scialom, Speckbacher, Mihaylov, Xiao, Karn, Goswami, Gupta, Ramanathan, Kerkez, Gonguet, Do, Vogeti, Albiero, Petrovic, Chu, Xiong, Fu, Meers, Martinet, Wang, Wang, Tan, Xia, Xie, Jia, Wang, Goldschlag, Gaur, Babaei, Wen, Song, Zhang, Li, Mao, Coudert, Yan, Chen, Papakipos, Singh, Srivastava, Jain, Kelsey, Shajnfeld, Gangidi, Victoria, Goldstand, Menon, Sharma, Boesenberg,
  Baevski, Feinstein, Kallet, Sangani, Teo, Yunus, Lupu, Alvarado, Caples, Gu, Ho, Poulton, Ryan, Ramchandani, Dong, Franco, Goyal, Saraf, Chowdhury, Gabriel, Bharambe, Eisenman, Yazdan, James, Maurer, Leonhardi, Huang, Loyd, Paola, Paranjape, Liu, Wu, Ni, Hancock, Wasti, Spence, Stojkovic, Gamido, Montalvo, Parker, Burton, Mejia, Liu, Wang, Kim, Zhou, Hu, Chu, Cai, Tindal, Feichtenhofer, Gao, Civin, Beaty, Kreymer, Li, Adkins, Xu, Testuggine, David, Parikh, Liskovich, Foss, Wang, Le, Holland, Dowling, Jamil, Montgomery, Presani, Hahn, Wood, Le, Brinkman, Arcaute, Dunbar, Smothers, Sun, Kreuk, Tian, Kokkinos, Ozgenel, Caggioni, Kanayet, Seide, Florez, Schwarz, Badeer, Swee, Halpern, Herman, Sizov, Guangyi, Zhang, Lakshminarayanan, Inan, Shojanazeri, Zou, Wang, Zha, Habeeb, Rudolph, Suk, Aspegren, Goldman, Zhan, Damlaj, Molybog, Tufanov, Leontiadis, Veliche, Gat, Weissman, Geboski, Kohli, Lam, Asher, Gaya, Marcus, Tang, Chan, Zhen, Reizenstein, Teboul, Zhong, Jin, Yang, Cummings, Carvill, Shepard, McPhie,
  Torres, Ginsburg, Wang, Wu, U, Saxena, Khandelwal, Zand, Matosich, Veeraraghavan, Michelena, Li, Jagadeesh, Huang, Chawla, Huang, Chen, Garg, A, Silva, Bell, Zhang, Guo, Yu, Moshkovich, Wehrstedt, Khabsa, Avalani, Bhatt, Mankus, Hasson, Lennie, Reso, Groshev, Naumov, Lathi, Keneally, Liu, Seltzer, Valko, Restrepo, Patel, Vyatskov, Samvelyan, Clark, Macey, Wang, Hermoso, Metanat, Rastegari, Bansal, Santhanam, Parks, White, Bawa, Singhal, Egebo, Usunier, Mehta, Laptev, Dong, Cheng, Chernoguz, Hart, Salpekar, Kalinli, Kent, Parekh, Saab, Balaji, Rittner, Bontrager, Roux, Dollar, Zvyagina, Ratanchandani, Yuvraj, Liang, Alao, Rodriguez, Ayub, Murthy, Nayani, Mitra, Parthasarathy, Li, Hogan, Battey, Wang, Howes, Rinott, Mehta, Siby, Bondu, Datta, Chugh, Hunt, Dhillon, Sidorov, Pan, Mahajan, Verma, Yamamoto, Ramaswamy, Lindsay, Lindsay, Feng, Lin, Zha, Patil, Shankar, Zhang, Zhang, Wang, Agarwal, Sajuyigbe, Chintala, Max, Chen, Kehoe, Satterfield, Govindaprasad, Gupta, Deng, Cho, Virk, Subramanian, Choudhury,
  Goldman, Remez, Glaser, Best, Koehler, Robinson, Li, Zhang, Matthews, Chou, Shaked, Vontimitta, Ajayi, Montanez, Mohan, Kumar, Mangla, Ionescu, Poenaru, Mihailescu, Ivanov, Li, Wang, Jiang, Bouaziz, Constable, Tang, Wu, Wang, Wu, Gao, Kleinman, Chen, Hu, Jia, Qi, Li, Zhang, Zhang, Adi, Nam, Yu, Wang, Zhao, Hao, Qian, Li, He, Rait, DeVito, Rosnbrick, Wen, Yang, Zhao, and Ma}]{grattafiori2024llama3herdmodels}
Aaron Grattafiori, Abhimanyu Dubey, Abhinav Jauhri, Abhinav Pandey, Abhishek Kadian, Ahmad Al-Dahle, Aiesha Letman, Akhil Mathur, Alan Schelten, Alex Vaughan, Amy Yang, Angela Fan, Anirudh Goyal, Anthony Hartshorn, Aobo Yang, Archi Mitra, Archie Sravankumar, Artem Korenev, Arthur Hinsvark, Arun Rao, Aston Zhang, Aurelien Rodriguez, Austen Gregerson, Ava Spataru, Baptiste Roziere, Bethany Biron, Binh Tang, Bobbie Chern, Charlotte Caucheteux, Chaya Nayak, Chloe Bi, Chris Marra, Chris McConnell, Christian Keller, Christophe Touret, Chunyang Wu, Corinne Wong, Cristian~Canton Ferrer, Cyrus Nikolaidis, Damien Allonsius, Daniel Song, Danielle Pintz, Danny Livshits, Danny Wyatt, David Esiobu, Dhruv Choudhary, Dhruv Mahajan, Diego Garcia-Olano, Diego Perino, Dieuwke Hupkes, Egor Lakomkin, Ehab AlBadawy, Elina Lobanova, Emily Dinan, Eric~Michael Smith, Filip Radenovic, Francisco Guzmán, Frank Zhang, Gabriel Synnaeve, Gabrielle Lee, Georgia~Lewis Anderson, Govind Thattai, Graeme Nail, Gregoire Mialon, Guan Pang,
  Guillem Cucurell, Hailey Nguyen, Hannah Korevaar, Hu~Xu, Hugo Touvron, Iliyan Zarov, Imanol~Arrieta Ibarra, Isabel Kloumann, Ishan Misra, Ivan Evtimov, Jack Zhang, Jade Copet, Jaewon Lee, Jan Geffert, Jana Vranes, Jason Park, Jay Mahadeokar, Jeet Shah, Jelmer van~der Linde, Jennifer Billock, Jenny Hong, Jenya Lee, Jeremy Fu, Jianfeng Chi, Jianyu Huang, Jiawen Liu, Jie Wang, Jiecao Yu, Joanna Bitton, Joe Spisak, Jongsoo Park, Joseph Rocca, Joshua Johnstun, Joshua Saxe, Junteng Jia, Kalyan~Vasuden Alwala, Karthik Prasad, Kartikeya Upasani, Kate Plawiak, Ke~Li, Kenneth Heafield, Kevin Stone, Khalid El-Arini, Krithika Iyer, Kshitiz Malik, Kuenley Chiu, Kunal Bhalla, Kushal Lakhotia, Lauren Rantala-Yeary, Laurens van~der Maaten, Lawrence Chen, Liang Tan, Liz Jenkins, Louis Martin, Lovish Madaan, Lubo Malo, Lukas Blecher, Lukas Landzaat, Luke de~Oliveira, Madeline Muzzi, Mahesh Pasupuleti, Mannat Singh, Manohar Paluri, Marcin Kardas, Maria Tsimpoukelli, Mathew Oldham, Mathieu Rita, Maya Pavlova, Melanie Kambadur,
  Mike Lewis, Min Si, Mitesh~Kumar Singh, Mona Hassan, Naman Goyal, Narjes Torabi, Nikolay Bashlykov, Nikolay Bogoychev, Niladri Chatterji, Ning Zhang, Olivier Duchenne, Onur Çelebi, Patrick Alrassy, Pengchuan Zhang, Pengwei Li, Petar Vasic, Peter Weng, Prajjwal Bhargava, Pratik Dubal, Praveen Krishnan, Punit~Singh Koura, Puxin Xu, Qing He, Qingxiao Dong, Ragavan Srinivasan, Raj Ganapathy, Ramon Calderer, Ricardo~Silveira Cabral, Robert Stojnic, Roberta Raileanu, Rohan Maheswari, Rohit Girdhar, Rohit Patel, Romain Sauvestre, Ronnie Polidoro, Roshan Sumbaly, Ross Taylor, Ruan Silva, Rui Hou, Rui Wang, Saghar Hosseini, Sahana Chennabasappa, Sanjay Singh, Sean Bell, Seohyun~Sonia Kim, Sergey Edunov, Shaoliang Nie, Sharan Narang, Sharath Raparthy, Sheng Shen, Shengye Wan, Shruti Bhosale, Shun Zhang, Simon Vandenhende, Soumya Batra, Spencer Whitman, Sten Sootla, Stephane Collot, Suchin Gururangan, Sydney Borodinsky, Tamar Herman, Tara Fowler, Tarek Sheasha, Thomas Georgiou, Thomas Scialom, Tobias Speckbacher,
  Todor Mihaylov, Tong Xiao, Ujjwal Karn, Vedanuj Goswami, Vibhor Gupta, Vignesh Ramanathan, Viktor Kerkez, Vincent Gonguet, Virginie Do, Vish Vogeti, Vítor Albiero, Vladan Petrovic, Weiwei Chu, Wenhan Xiong, Wenyin Fu, Whitney Meers, Xavier Martinet, Xiaodong Wang, Xiaofang Wang, Xiaoqing~Ellen Tan, Xide Xia, Xinfeng Xie, Xuchao Jia, Xuewei Wang, Yaelle Goldschlag, Yashesh Gaur, Yasmine Babaei, Yi~Wen, Yiwen Song, Yuchen Zhang, Yue Li, Yuning Mao, Zacharie~Delpierre Coudert, Zheng Yan, Zhengxing Chen, Zoe Papakipos, Aaditya Singh, Aayushi Srivastava, Abha Jain, Adam Kelsey, Adam Shajnfeld, Adithya Gangidi, Adolfo Victoria, Ahuva Goldstand, Ajay Menon, Ajay Sharma, Alex Boesenberg, Alexei Baevski, Allie Feinstein, Amanda Kallet, Amit Sangani, Amos Teo, Anam Yunus, Andrei Lupu, Andres Alvarado, Andrew Caples, Andrew Gu, Andrew Ho, Andrew Poulton, Andrew Ryan, Ankit Ramchandani, Annie Dong, Annie Franco, Anuj Goyal, Aparajita Saraf, Arkabandhu Chowdhury, Ashley Gabriel, Ashwin Bharambe, Assaf Eisenman, Azadeh
  Yazdan, Beau James, Ben Maurer, Benjamin Leonhardi, Bernie Huang, Beth Loyd, Beto~De Paola, Bhargavi Paranjape, Bing Liu, Bo~Wu, Boyu Ni, Braden Hancock, Bram Wasti, Brandon Spence, Brani Stojkovic, Brian Gamido, Britt Montalvo, Carl Parker, Carly Burton, Catalina Mejia, Ce~Liu, Changhan Wang, Changkyu Kim, Chao Zhou, Chester Hu, Ching-Hsiang Chu, Chris Cai, Chris Tindal, Christoph Feichtenhofer, Cynthia Gao, Damon Civin, Dana Beaty, Daniel Kreymer, Daniel Li, David Adkins, David Xu, Davide Testuggine, Delia David, Devi Parikh, Diana Liskovich, Didem Foss, Dingkang Wang, Duc Le, Dustin Holland, Edward Dowling, Eissa Jamil, Elaine Montgomery, Eleonora Presani, Emily Hahn, Emily Wood, Eric-Tuan Le, Erik Brinkman, Esteban Arcaute, Evan Dunbar, Evan Smothers, Fei Sun, Felix Kreuk, Feng Tian, Filippos Kokkinos, Firat Ozgenel, Francesco Caggioni, Frank Kanayet, Frank Seide, Gabriela~Medina Florez, Gabriella Schwarz, Gada Badeer, Georgia Swee, Gil Halpern, Grant Herman, Grigory Sizov, Guangyi, Zhang, Guna
  Lakshminarayanan, Hakan Inan, Hamid Shojanazeri, Han Zou, Hannah Wang, Hanwen Zha, Haroun Habeeb, Harrison Rudolph, Helen Suk, Henry Aspegren, Hunter Goldman, Hongyuan Zhan, Ibrahim Damlaj, Igor Molybog, Igor Tufanov, Ilias Leontiadis, Irina-Elena Veliche, Itai Gat, Jake Weissman, James Geboski, James Kohli, Janice Lam, Japhet Asher, Jean-Baptiste Gaya, Jeff Marcus, Jeff Tang, Jennifer Chan, Jenny Zhen, Jeremy Reizenstein, Jeremy Teboul, Jessica Zhong, Jian Jin, Jingyi Yang, Joe Cummings, Jon Carvill, Jon Shepard, Jonathan McPhie, Jonathan Torres, Josh Ginsburg, Junjie Wang, Kai Wu, Kam~Hou U, Karan Saxena, Kartikay Khandelwal, Katayoun Zand, Kathy Matosich, Kaushik Veeraraghavan, Kelly Michelena, Keqian Li, Kiran Jagadeesh, Kun Huang, Kunal Chawla, Kyle Huang, Lailin Chen, Lakshya Garg, Lavender A, Leandro Silva, Lee Bell, Lei Zhang, Liangpeng Guo, Licheng Yu, Liron Moshkovich, Luca Wehrstedt, Madian Khabsa, Manav Avalani, Manish Bhatt, Martynas Mankus, Matan Hasson, Matthew Lennie, Matthias Reso, Maxim
  Groshev, Maxim Naumov, Maya Lathi, Meghan Keneally, Miao Liu, Michael~L. Seltzer, Michal Valko, Michelle Restrepo, Mihir Patel, Mik Vyatskov, Mikayel Samvelyan, Mike Clark, Mike Macey, Mike Wang, Miquel~Jubert Hermoso, Mo~Metanat, Mohammad Rastegari, Munish Bansal, Nandhini Santhanam, Natascha Parks, Natasha White, Navyata Bawa, Nayan Singhal, Nick Egebo, Nicolas Usunier, Nikhil Mehta, Nikolay~Pavlovich Laptev, Ning Dong, Norman Cheng, Oleg Chernoguz, Olivia Hart, Omkar Salpekar, Ozlem Kalinli, Parkin Kent, Parth Parekh, Paul Saab, Pavan Balaji, Pedro Rittner, Philip Bontrager, Pierre Roux, Piotr Dollar, Polina Zvyagina, Prashant Ratanchandani, Pritish Yuvraj, Qian Liang, Rachad Alao, Rachel Rodriguez, Rafi Ayub, Raghotham Murthy, Raghu Nayani, Rahul Mitra, Rangaprabhu Parthasarathy, Raymond Li, Rebekkah Hogan, Robin Battey, Rocky Wang, Russ Howes, Ruty Rinott, Sachin Mehta, Sachin Siby, Sai~Jayesh Bondu, Samyak Datta, Sara Chugh, Sara Hunt, Sargun Dhillon, Sasha Sidorov, Satadru Pan, Saurabh Mahajan,
  Saurabh Verma, Seiji Yamamoto, Sharadh Ramaswamy, Shaun Lindsay, Shaun Lindsay, Sheng Feng, Shenghao Lin, Shengxin~Cindy Zha, Shishir Patil, Shiva Shankar, Shuqiang Zhang, Shuqiang Zhang, Sinong Wang, Sneha Agarwal, Soji Sajuyigbe, Soumith Chintala, Stephanie Max, Stephen Chen, Steve Kehoe, Steve Satterfield, Sudarshan Govindaprasad, Sumit Gupta, Summer Deng, Sungmin Cho, Sunny Virk, Suraj Subramanian, Sy~Choudhury, Sydney Goldman, Tal Remez, Tamar Glaser, Tamara Best, Thilo Koehler, Thomas Robinson, Tianhe Li, Tianjun Zhang, Tim Matthews, Timothy Chou, Tzook Shaked, Varun Vontimitta, Victoria Ajayi, Victoria Montanez, Vijai Mohan, Vinay~Satish Kumar, Vishal Mangla, Vlad Ionescu, Vlad Poenaru, Vlad~Tiberiu Mihailescu, Vladimir Ivanov, Wei Li, Wenchen Wang, Wenwen Jiang, Wes Bouaziz, Will Constable, Xiaocheng Tang, Xiaojian Wu, Xiaolan Wang, Xilun Wu, Xinbo Gao, Yaniv Kleinman, Yanjun Chen, Ye~Hu, Ye~Jia, Ye~Qi, Yenda Li, Yilin Zhang, Ying Zhang, Yossi Adi, Youngjin Nam, Yu, Wang, Yu~Zhao, Yuchen Hao, Yundi
  Qian, Yunlu Li, Yuzi He, Zach Rait, Zachary DeVito, Zef Rosnbrick, Zhaoduo Wen, Zhenyu Yang, Zhiwei Zhao, and Zhiyu Ma. 2024.
\newblock \href {https://arxiv.org/abs/2407.21783} {The llama 3 herd of models}.
\newblock \emph{Preprint}, arXiv:2407.21783.

\bibitem[{Gurgurov et~al.(2024)Gurgurov, Hartmann, and Ostermann}]{gurgurov-etal-2024-adapting}
Daniil Gurgurov, Mareike Hartmann, and Simon Ostermann. 2024.
\newblock \href {https://doi.org/10.18653/v1/2024.kallm-1.7} {Adapting multilingual {LLM}s to low-resource languages with knowledge graphs via adapters}.
\newblock In \emph{Proceedings of the 1st Workshop on Knowledge Graphs and Large Language Models (KaLLM 2024)}, pages 63--74, Bangkok, Thailand. Association for Computational Linguistics.

\bibitem[{Hardalov et~al.(2020)Hardalov, Mihaylov, Zlatkova, Dinkov, Koychev, and Nakov}]{exams}
Momchil Hardalov, Todor Mihaylov, Dimitrina Zlatkova, Yoan Dinkov, Ivan Koychev, and Preslav Nakov. 2020.
\newblock \href {https://doi.org/10.18653/v1/2020.emnlp-main.438} {{EXAMS}: A multi-subject high school examinations dataset for cross-lingual and multilingual question answering}.
\newblock In \emph{Proceedings of the 2020 Conference on Empirical Methods in Natural Language Processing (EMNLP)}, pages 5427--5444, Online. Association for Computational Linguistics.

\bibitem[{He et~al.(2024)He, Benhaim, Patra, Vaddamanu, Ahuja, Chopra, Chaudhary, Zhao, and Song}]{he2024scalinglawsmultilinguallanguage}
Yifei He, Alon Benhaim, Barun Patra, Praneetha Vaddamanu, Sanchit Ahuja, Parul Chopra, Vishrav Chaudhary, Han Zhao, and Xia Song. 2024.
\newblock \href {https://arxiv.org/abs/2410.12883} {Scaling laws for multilingual language models}.
\newblock \emph{Preprint}, arXiv:2410.12883.

\bibitem[{Hendrycks et~al.(2021)Hendrycks, Burns, Basart, Zou, Mazeika, Song, and Steinhardt}]{mmlu}
Dan Hendrycks, Collin Burns, Steven Basart, Andy Zou, Mantas Mazeika, Dawn Song, and Jacob Steinhardt. 2021.
\newblock \href {https://openreview.net/forum?id=d7KBjmI3GmQ} {Measuring massive multitask language understanding}.
\newblock In \emph{International Conference on Learning Representations}.

\bibitem[{Ibrahim et~al.(2024)Ibrahim, Th{\'e}rien, Gupta, Richter, Anthony, Belilovsky, Lesort, and Rish}]{ibrahim2024simple}
Adam Ibrahim, Benjamin Th{\'e}rien, Kshitij Gupta, Mats~Leon Richter, Quentin~Gregory Anthony, Eugene Belilovsky, Timoth{\'e}e Lesort, and Irina Rish. 2024.
\newblock \href {https://openreview.net/forum?id=DimPeeCxKO} {Simple and scalable strategies to continually pre-train large language models}.
\newblock \emph{Transactions on Machine Learning Research}.

\bibitem[{Jiang et~al.(2023)Jiang, Sablayrolles, Mensch, Bamford, Chaplot, de~las Casas, Bressand, Lengyel, Lample, Saulnier, Lavaud, Lachaux, Stock, Scao, Lavril, Wang, Lacroix, and Sayed}]{jiang2023mistral7b}
Albert~Q. Jiang, Alexandre Sablayrolles, Arthur Mensch, Chris Bamford, Devendra~Singh Chaplot, Diego de~las Casas, Florian Bressand, Gianna Lengyel, Guillaume Lample, Lucile Saulnier, Lélio~Renard Lavaud, Marie-Anne Lachaux, Pierre Stock, Teven~Le Scao, Thibaut Lavril, Thomas Wang, Timothée Lacroix, and William~El Sayed. 2023.
\newblock \href {https://arxiv.org/abs/2310.06825} {Mistral 7b}.
\newblock \emph{Preprint}, arXiv:2310.06825.

\bibitem[{Joshi et~al.(2017)Joshi, Choi, Weld, and Zettlemoyer}]{triviaqa}
Mandar Joshi, Eunsol Choi, Daniel Weld, and Luke Zettlemoyer. 2017.
\newblock \href {https://doi.org/10.18653/v1/P17-1147} {{T}rivia{QA}: A large scale distantly supervised challenge dataset for reading comprehension}.
\newblock In \emph{Proceedings of the 55th Annual Meeting of the Association for Computational Linguistics (Volume 1: Long Papers)}, pages 1601--1611, Vancouver, Canada. Association for Computational Linguistics.

\bibitem[{Joshi et~al.(2024)Joshi, Singla, Kamath, Kalani, Paul, Vaidya, Chauhan, Wartikar, and Long}]{joshi2024adaptingmultilingualllmslowresource}
Raviraj Joshi, Kanishk Singla, Anusha Kamath, Raunak Kalani, Rakesh Paul, Utkarsh Vaidya, Sanjay~Singh Chauhan, Niranjan Wartikar, and Eileen Long. 2024.
\newblock \href {https://arxiv.org/abs/2410.14815} {Adapting multilingual llms to low-resource languages using continued pre-training and synthetic corpus}.
\newblock \emph{Preprint}, arXiv:2410.14815.

\bibitem[{Lin et~al.(2024)Lin, Deng, Chandu, Brahman, Ravichander, Pyatkin, Dziri, Bras, and Choi}]{lin2024wildbench}
Bill~Yuchen Lin, Yuntian Deng, Khyathi Chandu, Faeze Brahman, Abhilasha Ravichander, Valentina Pyatkin, Nouha Dziri, Ronan~Le Bras, and Yejin Choi. 2024.
\newblock Wildbench: Benchmarking llms with challenging tasks from real users in the wild.
\newblock \emph{arXiv preprint arXiv:2406.04770}.

\bibitem[{Luukkonen et~al.(2023)Luukkonen, Komulainen, Luoma, Eskelinen, Kanerva, Kupari, Ginter, Laippala, Muennighoff, Piktus, Wang, Tazi, Scao, Wolf, Suominen, Sairanen, Merioksa, Heinonen, Vahtola, Antao, and Pyysalo}]{luukkonen-etal-2023-fingpt}
Risto Luukkonen, Ville Komulainen, Jouni Luoma, Anni Eskelinen, Jenna Kanerva, Hanna-Mari Kupari, Filip Ginter, Veronika Laippala, Niklas Muennighoff, Aleksandra Piktus, Thomas Wang, Nouamane Tazi, Teven Scao, Thomas Wolf, Osma Suominen, Samuli Sairanen, Mikko Merioksa, Jyrki Heinonen, Aija Vahtola, Samuel Antao, and Sampo Pyysalo. 2023.
\newblock \href {https://doi.org/10.18653/v1/2023.emnlp-main.164} {{F}in{GPT}: Large generative models for a small language}.
\newblock In \emph{Proceedings of the 2023 Conference on Empirical Methods in Natural Language Processing}, pages 2710--2726, Singapore. Association for Computational Linguistics.

\bibitem[{Marchisio et~al.(2024)Marchisio, Ko, Bérard, Dehaze, and Ruder}]{languageconfusion}
Kelly Marchisio, Wei-Yin Ko, Alexandre Bérard, Théo Dehaze, and Sebastian Ruder. 2024.
\newblock \href {https://arxiv.org/abs/2406.20052} {Understanding and mitigating language confusion in llms}.
\newblock \emph{Preprint}, arXiv:2406.20052.

\bibitem[{Masala et~al.(2024)Masala, Ilie-Ablachim, Dima, Corlatescu, Zavelca, Olaru, Terian, Terian, Leordeanu, Velicu, Popescu, Dascalu, and Rebedea}]{OpenLLMRomanian}
Mihai Masala, Denis~C. Ilie-Ablachim, Alexandru Dima, Dragos Corlatescu, Miruna Zavelca, Ovio Olaru, Simina Terian, Andrei Terian, Marius Leordeanu, Horia Velicu, Marius Popescu, Mihai Dascalu, and Traian Rebedea. 2024.
\newblock \href {https://arxiv.org/abs/2406.18266} {"vorbe\c{s}ti rom\^ane\c{s}te?" a recipe to train powerful romanian llms with english instructions}.
\newblock \emph{Preprint}, arXiv:2406.18266.

\bibitem[{Nguyen et~al.(2023)Nguyen, Pham, and Dao}]{vinallamallamabasedvietnamesefoundation}
Quan Nguyen, Huy Pham, and Dung Dao. 2023.
\newblock \href {https://arxiv.org/abs/2312.11011} {Vinallama: Llama-based vietnamese foundation model}.
\newblock \emph{Preprint}, arXiv:2312.11011.

\bibitem[{Nikolich et~al.(2024)Nikolich, Korolev, Bratchikov, Kiselev, and Shelmanov}]{vikhr}
Aleksandr Nikolich, Konstantin Korolev, Sergei Bratchikov, Igor Kiselev, and Artem Shelmanov. 2024.
\newblock \href {https://doi.org/10.18653/v1/2024.mrl-1.15} {Vikhr: Constructing a state-of-the-art bilingual open-source instruction-following large language model for {R}ussian}.
\newblock In \emph{Proceedings of the Fourth Workshop on Multilingual Representation Learning (MRL 2024)}, pages 189--199, Miami, Florida, USA. Association for Computational Linguistics.

\bibitem[{of~Bulgaria(2024)}]{mon}
Republic of~Bulgaria. 2024.
\newblock \href {https://mon.bg/} {Ministry of education and science}.

\bibitem[{Paszke et~al.(2019)Paszke, Gross, Massa, Lerer, Bradbury, Chanan, Killeen, Lin, Gimelshein, Antiga, Desmaison, K{\"{o}}pf, Yang, DeVito, Raison, Tejani, Chilamkurthy, Steiner, Fang, Bai, and Chintala}]{PaszkeGMLBCKLGA19}
Adam Paszke, Sam Gross, Francisco Massa, Adam Lerer, James Bradbury, Gregory Chanan, Trevor Killeen, Zeming Lin, Natalia Gimelshein, Luca Antiga, Alban Desmaison, Andreas K{\"{o}}pf, Edward~Z. Yang, Zachary DeVito, Martin Raison, Alykhan Tejani, Sasank Chilamkurthy, Benoit Steiner, Lu~Fang, Junjie Bai, and Soumith Chintala. 2019.
\newblock \href {https://proceedings.neurips.cc/paper/2019/hash/bdbca288fee7f92f2bfa9f7012727740-Abstract.html} {Pytorch: An imperative style, high-performance deep learning library}.
\newblock In \emph{Advances in Neural Information Processing Systems 32: Annual Conference on Neural Information Processing Systems 2019, NeurIPS 2019, December 8-14, 2019, Vancouver, BC, Canada}, pages 8024--8035.

\bibitem[{Pl{\"u}ster(2023)}]{LeoLM}
Bj{\"o}rn Pl{\"u}ster. 2023.
\newblock \href {https://laion.ai/blog/leo-lm/} {Leolm: Igniting german-language llm research}.

\bibitem[{Radford et~al.(2019)Radford, Wu, Child, Luan, Amodei, and Sutskever}]{Radford2019LanguageMA}
Alec Radford, Jeff Wu, Rewon Child, David Luan, Dario Amodei, and Ilya Sutskever. 2019.
\newblock \href {https://api.semanticscholar.org/CorpusID:160025533} {Language models are unsupervised multitask learners}.

\bibitem[{Rafailov et~al.(2023)Rafailov, Sharma, Mitchell, Manning, Ermon, and Finn}]{rafailov2023direct}
Rafael Rafailov, Archit Sharma, Eric Mitchell, Christopher~D Manning, Stefano Ermon, and Chelsea Finn. 2023.
\newblock \href {https://openreview.net/forum?id=HPuSIXJaa9} {Direct preference optimization: Your language model is secretly a reward model}.
\newblock In \emph{Thirty-seventh Conference on Neural Information Processing Systems}.

\bibitem[{Rajbhandari et~al.(2020)Rajbhandari, Rasley, Ruwase, and He}]{zero9355301}
Samyam Rajbhandari, Jeff Rasley, Olatunji Ruwase, and Yuxiong He. 2020.
\newblock \href {https://doi.org/10.1109/SC41405.2020.00024} {Zero: Memory optimizations toward training trillion parameter models}.
\newblock In \emph{SC20: International Conference for High Performance Computing, Networking, Storage and Analysis}, pages 1--16.

\bibitem[{Rasley et~al.(2020)Rasley, Rajbhandari, Ruwase, and He}]{deepspeed10.1145/3394486.3406703}
Jeff Rasley, Samyam Rajbhandari, Olatunji Ruwase, and Yuxiong He. 2020.
\newblock \href {https://doi.org/10.1145/3394486.3406703} {Deepspeed: System optimizations enable training deep learning models with over 100 billion parameters}.
\newblock In \emph{Proceedings of the 26th ACM SIGKDD International Conference on Knowledge Discovery \& Data Mining}, KDD '20, page 3505–3506, New York, NY, USA. Association for Computing Machinery.

\bibitem[{Sakaguchi et~al.(2021)Sakaguchi, Bras, Bhagavatula, and Choi}]{winogrande}
Keisuke Sakaguchi, Ronan~Le Bras, Chandra Bhagavatula, and Yejin Choi. 2021.
\newblock \href {https://doi.org/10.1145/3474381} {Winogrande: an adversarial winograd schema challenge at scale}.
\newblock \emph{Commun. ACM}, 64(9):99–106.

\bibitem[{Scialom et~al.(2022)Scialom, Chakrabarty, and Muresan}]{ScialomCM22}
Thomas Scialom, Tuhin Chakrabarty, and Smaranda Muresan. 2022.
\newblock \href {https://doi.org/10.18653/V1/2022.EMNLP-MAIN.410} {Fine-tuned language models are continual learners}.
\newblock In \emph{Proceedings of the 2022 Conference on Empirical Methods in Natural Language Processing, {EMNLP} 2022, Abu Dhabi, United Arab Emirates, December 7-11, 2022}, pages 6107--6122. Association for Computational Linguistics.

\bibitem[{Shi et~al.(2024)Shi, Xu, Wang, Qin, Wang, Wang, and Wang}]{ShiXWQWWW24}
Haizhou Shi, Zihao Xu, Hengyi Wang, Weiyi Qin, Wenyuan Wang, Yibin Wang, and Hao Wang. 2024.
\newblock \href {https://doi.org/10.48550/ARXIV.2404.16789} {Continual learning of large language models: {A} comprehensive survey}.
\newblock \emph{CoRR}, abs/2404.16789.

\bibitem[{Team(2024)}]{qwen2.5}
Qwen Team. 2024.
\newblock \href {https://qwenlm.github.io/blog/qwen2.5/} {Qwen2.5: A party of foundation models}.

\bibitem[{Toraman(2024)}]{turkish}
Cagri Toraman. 2024.
\newblock \href {https://doi.org/10.18653/v1/2024.mrl-1.3} {Adapting open-source generative large language models for low-resource languages: A case study for {T}urkish}.
\newblock In \emph{Proceedings of the Fourth Workshop on Multilingual Representation Learning (MRL 2024)}, pages 30--44, Miami, Florida, USA. Association for Computational Linguistics.

\bibitem[{Touvron et~al.(2023{\natexlab{a}})Touvron, Lavril, Izacard, Martinet, Lachaux, Lacroix, Rozière, Goyal, Hambro, Azhar, Rodriguez, Joulin, Grave, and Lample}]{touvron2023llamaopenefficientfoundation}
Hugo Touvron, Thibaut Lavril, Gautier Izacard, Xavier Martinet, Marie-Anne Lachaux, Timothée Lacroix, Baptiste Rozière, Naman Goyal, Eric Hambro, Faisal Azhar, Aurelien Rodriguez, Armand Joulin, Edouard Grave, and Guillaume Lample. 2023{\natexlab{a}}.
\newblock \href {https://arxiv.org/abs/2302.13971} {Llama: Open and efficient foundation language models}.
\newblock \emph{Preprint}, arXiv:2302.13971.

\bibitem[{Touvron et~al.(2023{\natexlab{b}})Touvron, Martin, Stone, Albert, Almahairi, Babaei, Bashlykov, Batra, Bhargava, Bhosale, Bikel, Blecher, Ferrer, Chen, Cucurull, Esiobu, Fernandes, Fu, Fu, Fuller, Gao, Goswami, Goyal, Hartshorn, Hosseini, Hou, Inan, Kardas, Kerkez, Khabsa, Kloumann, Korenev, Koura, Lachaux, Lavril, Lee, Liskovich, Lu, Mao, Martinet, Mihaylov, Mishra, Molybog, Nie, Poulton, Reizenstein, Rungta, Saladi, Schelten, Silva, Smith, Subramanian, Tan, Tang, Taylor, Williams, Kuan, Xu, Yan, Zarov, Zhang, Fan, Kambadur, Narang, Rodriguez, Stojnic, Edunov, and Scialom}]{touvron2023llama}
Hugo Touvron, Louis Martin, Kevin Stone, Peter Albert, Amjad Almahairi, Yasmine Babaei, Nikolay Bashlykov, Soumya Batra, Prajjwal Bhargava, Shruti Bhosale, Dan Bikel, Lukas Blecher, Cristian~Canton Ferrer, Moya Chen, Guillem Cucurull, David Esiobu, Jude Fernandes, Jeremy Fu, Wenyin Fu, Brian Fuller, Cynthia Gao, Vedanuj Goswami, Naman Goyal, Anthony Hartshorn, Saghar Hosseini, Rui Hou, Hakan Inan, Marcin Kardas, Viktor Kerkez, Madian Khabsa, Isabel Kloumann, Artem Korenev, Punit~Singh Koura, Marie-Anne Lachaux, Thibaut Lavril, Jenya Lee, Diana Liskovich, Yinghai Lu, Yuning Mao, Xavier Martinet, Todor Mihaylov, Pushkar Mishra, Igor Molybog, Yixin Nie, Andrew Poulton, Jeremy Reizenstein, Rashi Rungta, Kalyan Saladi, Alan Schelten, Ruan Silva, Eric~Michael Smith, Ranjan Subramanian, Xiaoqing~Ellen Tan, Binh Tang, Ross Taylor, Adina Williams, Jian~Xiang Kuan, Puxin Xu, Zheng Yan, Iliyan Zarov, Yuchen Zhang, Angela Fan, Melanie Kambadur, Sharan Narang, Aurelien Rodriguez, Robert Stojnic, Sergey Edunov, and Thomas
  Scialom. 2023{\natexlab{b}}.
\newblock \href {https://arxiv.org/abs/2307.09288} {Llama 2: Open foundation and fine-tuned chat models}.
\newblock \emph{Preprint}, arXiv:2307.09288.

\bibitem[{Vo et~al.(2024)Vo, Jung, Lee, and Choi}]{vo2024redwhaleadaptedkoreanllm}
Anh-Dung Vo, Minseong Jung, Wonbeen Lee, and Daewoo Choi. 2024.
\newblock \href {https://arxiv.org/abs/2408.11294} {Redwhale: An adapted korean llm through efficient continual pretraining}.
\newblock \emph{Preprint}, arXiv:2408.11294.

\bibitem[{Wang et~al.(2024)Wang, Minervini, and Ponti}]{wang-etal-2024-probing-emergence}
Hetong Wang, Pasquale Minervini, and Edoardo Ponti. 2024.
\newblock \href {https://doi.org/10.18653/v1/2024.findings-acl.724} {Probing the emergence of cross-lingual alignment during {LLM} training}.
\newblock In \emph{Findings of the Association for Computational Linguistics: ACL 2024}, pages 12159--12173, Bangkok, Thailand. Association for Computational Linguistics.

\bibitem[{Weber et~al.(2024)Weber, Fu, Anthony, Oren, Adams, Alexandrov, Lyu, Nguyen, Yao, Adams, Athiwaratkun, Chalamala, Chen, Ryabinin, Dao, Liang, Re, Rish, and Zhang}]{weber2024redpajama}
Maurice Weber, Daniel~Y Fu, Quentin~Gregory Anthony, Yonatan Oren, Shane Adams, Anton Alexandrov, Xiaozhong Lyu, Huu Nguyen, Xiaozhe Yao, Virginia Adams, Ben Athiwaratkun, Rahul Chalamala, Kezhen Chen, Max Ryabinin, Tri Dao, Percy Liang, Christopher Re, Irina Rish, and Ce~Zhang. 2024.
\newblock \href {https://openreview.net/forum?id=lnuXaRpwvw} {Redpajama: an open dataset for training large language models}.
\newblock In \emph{The Thirty-eight Conference on Neural Information Processing Systems Datasets and Benchmarks Track}.

\bibitem[{Wolf et~al.(2019)Wolf, Debut, Sanh, Chaumond, Delangue, Moi, Cistac, Rault, Louf, Funtowicz, and Brew}]{WolfDSCD19}
Thomas Wolf, Lysandre Debut, Victor Sanh, Julien Chaumond, Clement Delangue, Anthony Moi, Pierric Cistac, Tim Rault, R{\'{e}}mi Louf, Morgan Funtowicz, and Jamie Brew. 2019.
\newblock \href {https://arxiv.org/abs/1910.03771} {Huggingface's transformers: State-of-the-art natural language processing}.
\newblock \emph{CoRR}, abs/1910.03771.

\bibitem[{Yang et~al.(2024)Yang, Yang, Hui, Zheng, Yu, Zhou, Li, Li, Liu, Huang et~al.}]{qwen2}
An~Yang, Baosong Yang, Binyuan Hui, Bo~Zheng, Bowen Yu, Chang Zhou, Chengpeng Li, Chengyuan Li, Dayiheng Liu, Fei Huang, et~al. 2024.
\newblock Qwen2 technical report.
\newblock \emph{arXiv preprint arXiv:2407.10671}.

\bibitem[{Zellers et~al.(2019)Zellers, Holtzman, Bisk, Farhadi, and Choi}]{hellaswag}
Rowan Zellers, Ari Holtzman, Yonatan Bisk, Ali Farhadi, and Yejin Choi. 2019.
\newblock \href {https://doi.org/10.18653/v1/P19-1472} {{H}ella{S}wag: Can a machine really finish your sentence?}
\newblock In \emph{Proceedings of the 57th Annual Meeting of the Association for Computational Linguistics}, pages 4791--4800, Florence, Italy. Association for Computational Linguistics.

\bibitem[{Zhang et~al.(2023)Zhang, Fang, Chen, and Namazi{-}Rad}]{ZhangF0N23}
Zihan Zhang, Meng Fang, Ling Chen, and Mohammad{-}Reza Namazi{-}Rad. 2023.
\newblock \href {https://doi.org/10.18653/V1/2023.FINDINGS-EMNLP.633} {{CITB:} {A} benchmark for continual instruction tuning}.
\newblock In \emph{Findings of the Association for Computational Linguistics: {EMNLP} 2023, Singapore, December 6-10, 2023}, pages 9443--9455. Association for Computational Linguistics.

\bibitem[{Zhao et~al.(2024)Zhao, Zhang, Gao, Zhang, Gui, and Huang}]{llamaenglishempiricalstudy}
Jun Zhao, Zhihao Zhang, Luhui Gao, Qi~Zhang, Tao Gui, and Xuanjing Huang. 2024.
\newblock \href {https://arxiv.org/abs/2401.01055} {Llama beyond english: An empirical study on language capability transfer}.
\newblock \emph{Preprint}, arXiv:2401.01055.

\bibitem[{Zheng et~al.(2024)Zheng, Chiang, Sheng, Zhuang, Wu, Zhuang, Lin, Li, Li, Xing, Zhang, Gonzalez, and Stoica}]{ChatbotArena2023}
Lianmin Zheng, Wei-Lin Chiang, Ying Sheng, Siyuan Zhuang, Zhanghao Wu, Yonghao Zhuang, Zi~Lin, Zhuohan Li, Dacheng Li, Eric~P. Xing, Hao Zhang, Joseph~E. Gonzalez, and Ion Stoica. 2024.
\newblock Judging llm-as-a-judge with mt-bench and chatbot arena.
\newblock In \emph{Proceedings of the 37th International Conference on Neural Information Processing Systems}, NIPS '23, Red Hook, NY, USA. Curran Associates Inc.

\bibitem[{Zhuang et~al.(2020)Zhuang, Qi, Duan, Xi, Zhu, Zhu, Xiong, and He}]{zhuang2020comprehensivesurveytransferlearning}
Fuzhen Zhuang, Zhiyuan Qi, Keyu Duan, Dongbo Xi, Yongchun Zhu, Hengshu Zhu, Hui Xiong, and Qing He. 2020.
\newblock \href {https://arxiv.org/abs/1911.02685} {A comprehensive survey on transfer learning}.
\newblock \emph{Preprint}, arXiv:1911.02685.

\end{thebibliography}

\clearpage

\appendix

\begin{table*}[tbhp]
    \centering
    \caption{English scores on benchmark process of \modelbig starting from Gemma-2-27B. As we are not aiming particularly on English skills, this mostly measures the effects on forgetting by training the model on Bulgarian text.}
      \label{tab:forgetting}
    \resizebox{\linewidth}{!}{
    \begin{tabular}{lx{2}{2}x{2}{2}|x{2}{1}x{2}{1}x{2}{1}x{2}{1}x{2}{1}x{2}{1}x{2}{1}x{2}{1}}
        \toprule
        Model & {Bulgarian} & {English} & \multicolumn{8}{c}{Scores on original English benchmarks} \\
              & {(average)} & {(average)} & {Wino} & {Hellaswag} & {Arc-C} & {Arc-E} & {MMLU} & {MathQA} & {GSM-8K} & {TriviaQA} \\
        \midrule
        Gemma-2-27B & 60.95 & 74.47  &  80.6 & 84.9 & 65.6 & 88.3 & 75.5 & 48.9 & 73.9 & 78.0 \\
        \midrule
        Stage G1 & 62.43 & 74.15  &  79.4 & 83.3 & 65.4 & 88.3 & 74.4 & 48.2 & 77.5 & 76.7 \\
        Stage G2 & 60.57  & 73.36  &  79.6 & 84.7 & 61.7 & 84.6 & 74.6 & 47.2 & 80.2 & 74.2 \\
        Merge G1$\oplus$2 (slerp) & 62.53  & 74.76  &  80.3 & 83.8 & 65.6 & 88.5 & 74.5 & 47.9 & 80.0 & 77.4 \\
        \midrule
        Stage G3 & 62.43 & 73.99  &  80.4 & 83.5 & 64.2 & 87.5 & 74.7 & 47.7 & 77.5 & 76.3 \\
        Stage G4 & 62.01  & 73.25  &  79.2 & 84.5 & 61.4 & 83.7 & 74.6 & 47.9 & 80.2 & 74.4 \\
        Gemma-2-27-it (IT) & 60.51 & 75.08  &  78 & 84.2 & 66.9 & 83.6 & 76.2 & 48.6 & 85.9 & 77.2 \\
        Merge G4$\oplus$IT & 63.97 & 75.99  &  78.3 & 85.8 & 68.3 & 86.4 & 76.5 & 48.6 & 85.7 & 78.2 \\
        Merge G3$\oplus$(G4$\oplus$IT) & 64.59  & 75.89  &  79.5 & 85.1 & 67.7 & 88.2 & 76.1 & 48.4 & 83.9 & 78.3 \\
        \midrule
        Stage G5 & 63.03 & 73.84  &  79.2 & 83.9 & 62.8 & 86 & 75.4 & 46.6 & 80.4 & 76.4 \\
        Stage G6 & 62.88 & 73.39  &  78.7 & 84.9 & 60.7 & 84.9 & 74.9 & 46.7 & 81.7 & 74.8 \\
        G5$\oplus$G6 & 63.42  & 74.11  &  79.1 & 84.2 & 62.6 & 85.9 & 75.3 & 47.3 & 82 & 76.6 \\
        \midrule
        Stage G7 & 63.40 & 73.75  &  79 & 83.9 & 62.5 & 86 & 74.7 & 47.3 & 80.6 & 75.9 \\
        Stage G8 & 63.05 & 73.52  &  78.5 & 84.8 & 61.9 & 84.7 & 75 & 46.3 & 81.9 & 75.1 \\
        G8$\oplus$IT & 64.35 & 75.87  &  77.7 & 85.9 & 67.8 & 86.6 & 76.5 & 48.5 & 85.7 & 78.3 \\
        G7$\oplus$(G8$\oplus$IT) = \textbf{base} & 65.06 & 75.57  &  77.7 & 85.5 & 66.5 & 87.2 & 75.8 & 49 & 84.8 & 78.1 \\
        \midrule
        \modelbig= \textbf{base+IFT} (*) & 64.62 & 75.12  &  77.1 & 85 & 67.7 & 86.2 & 75.3 & 49.9 & 84.7 & 75.1 \\
        ((base+BG\_IFT)$\oplus$(base+EN\_IFT))$\oplus$(base+IFT) & 65.49 & 75.93 & 78.14 & 85.44 & 68.34 & 87.5 & 75.73 & 50.02 & 86.13 & 76.18 \\
        \bottomrule
        \end{tabular}
    }
\end{table*}

\begin{table*}[ht]
\small
\centering
\begin{tabularx}{\textwidth}{|P{3cm}|P{3cm}|P{4cm}|P{4cm}|}
\hline
\textbf{Question} & \textbf{Transformed Question} & \textbf{Answer} & \textbf{Judge Output} \\
\hline
\foreignlanguage{bulgarian}{Кой от авторите работи в епохата на Реализма?

А) "Сервантес"

Б) "Шекспир"

В) "Балзак"

Г) "Пушкин"} &
\foreignlanguage{bulgarian}{Кой известен писател е свързан с епохата на Реализма и е автор на "Човешка комедия"?} &
\foreignlanguage{bulgarian}{Известният френски писател, свързан с епохата на Реализма и автор на "Човешка комедия", е Оноре дьо Балзак. Той е един от най-значимите представители на реалистичната литература...} &
The answer provided is fully correct and written in excellent Bulgarian. It accurately identifies Honoré de Balzac as the French writer associated with the Realism era and the author of "La Comédie Humaine"...

6
 \\
\hline
\foreignlanguage{bulgarian}{	
В кой ред правилно е отразена валентността и степента на окисление на Фосфора във формулата на фосфорната киселина ($H_3PO_4$)?

А) 5; + 5

Б) 3; +3

В) 3; - 1

Г) 5; - 5
} &
\foreignlanguage{bulgarian}{Каква е валентността и степента на окисление на фосфора във формулата на фосфорната киселина ($H_3PO_4$)?} &
\foreignlanguage{bulgarian}{За да определим валентността и степента на окисление на фосфора в тази молекула, трябва първо да разберем структурата и връзките в нея... ...Обобщено:

- Валентността на фосфора във фосфорната киселина е 5.

- Степента на окисление на фосфора във фосфорната киселина е +5. } &
The provided answer is factually accurate, explains both the valency and oxidation state of phosphorus in ($H_3PO_4$) in detail, and uses proper and well-structured Bulgarian language without errors. It demonstrates a clear understanding of the topic and includes relevant calculations.

6
 \\
\hline
\foreignlanguage{bulgarian}{Какво се е случило през 870г.?

А) признаване на царската титла на Петър

Б) учредяването на Българската архиепископия

В) победата на хан Тервел над арабите

Г) присъединяване на Сердика към българската държава

} &
\foreignlanguage{bulgarian}{Какво значимо събитие, свързано с църковната история на България, се е случило през 870 година?} &
\foreignlanguage{bulgarian}{През 870 година се е случило значимо събитие, свързано с църковната история на България, а именно Покръстването на българите. Това събитие е от изключителна важност за българската история и култура, тъй като с него се поставя началото на християнизацията на българския народ и се утвърждава новата религия като официална в държавата...
} &
The provided answer is factually incorrect regarding the specific event in 870, which was the establishment of the Bulgarian Archbishopric, not the Christianization of Bulgaria... While the answer is well-written in Bulgarian and provides relevant historical context, it does not address the correct event for the year 870. Therefore, it shows some understanding of the broader historical context but fails to answer the specific question correctly.

3
\\
\hline
\foreignlanguage{bulgarian}{	
Вродените идеи са характерни за:

А) "Лок, Декарт, Хюм"

Б) "Декарт, Спиноза, Лок"

В) "Декарт, Лайбниц, Спиноза"

Г) "Лок, Бъркли, Хюм"

} &
\foreignlanguage{bulgarian}{Кои философи са известни с концепцията за вродените идеи?} &
\foreignlanguage{bulgarian}{Няколко философи са известни с концепцията за вродените идеи, включително Платон, Рене Декарт и Готфрид Лайбниц.} &
The answer correctly identifies two of the philosophers mentioned in the correct answer: René Descartes and Gottfried Leibniz. However, it incorrectly includes Plato, who is not part of the correct answer list. The language used is clear and grammatically correct. The inclusion of Plato is a minor error, but the answer still demonstrates a good understanding of the topic. Therefore, the answer is mostly correct with a minor omission.

5

\\
\hline
% Add more rows here
\end{tabularx}
\caption{Examples of rephrased multi-choice questions from the MON dataset with answers from various models that are then judged with given possible correct and incorrect answers.}
\label{tab:mon_examples}
\end{table*}

\begin{figure*}[htbp]
    \centering
    \begin{minipage}[t]{0.45\textwidth}
        \centering
        \includegraphics[width=\textwidth]{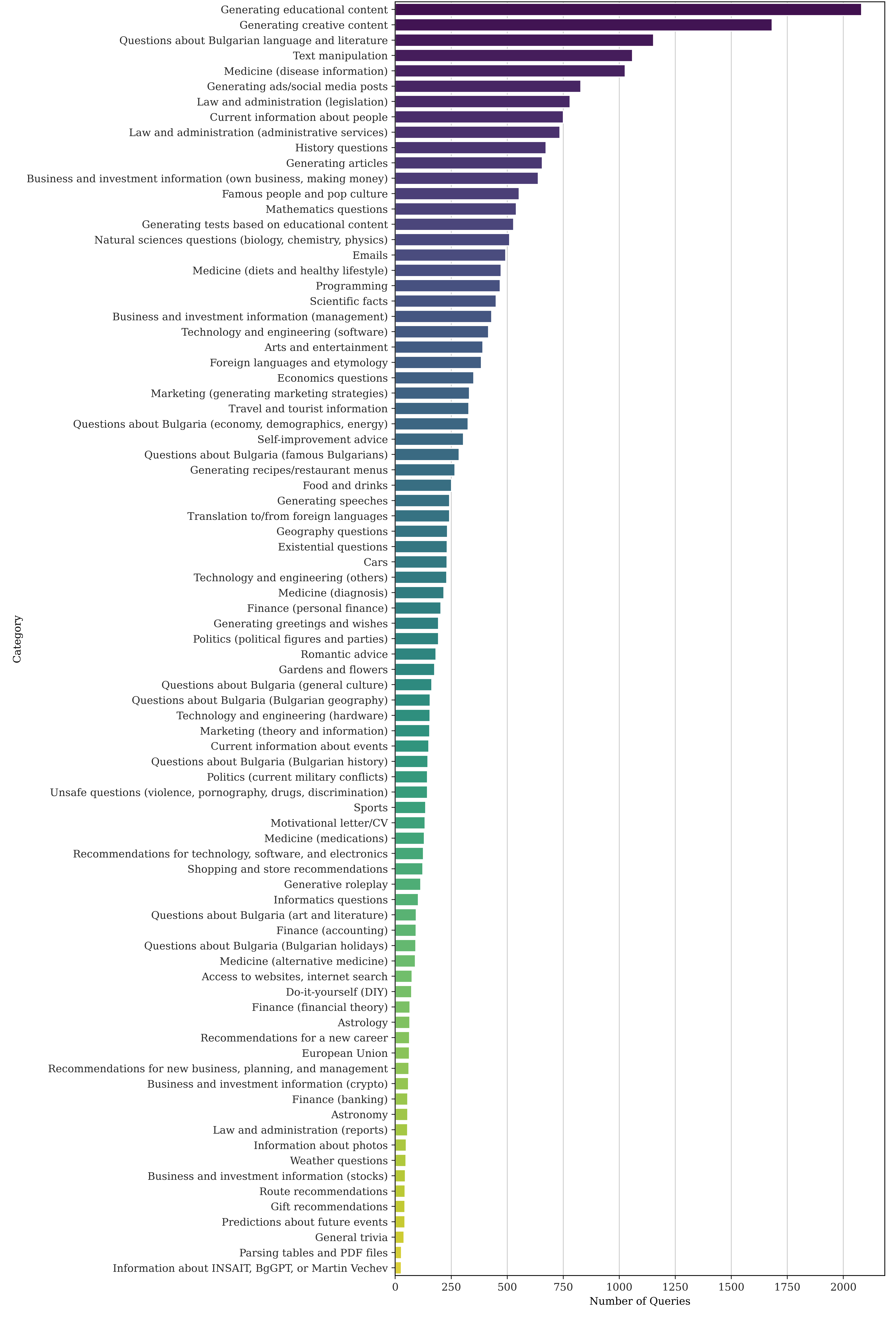} % Replace with your file
        \caption{Distribution of the 82 topics assigned as relevant to the queries from the chat IFT dataset. The category "Other" is removed from the visualization.}
        \label{fig:chat_topics_eval}
    \end{minipage}
    \hfill
    \begin{minipage}[t]{0.45\textwidth}
        \centering
        \includegraphics[width=\textwidth]{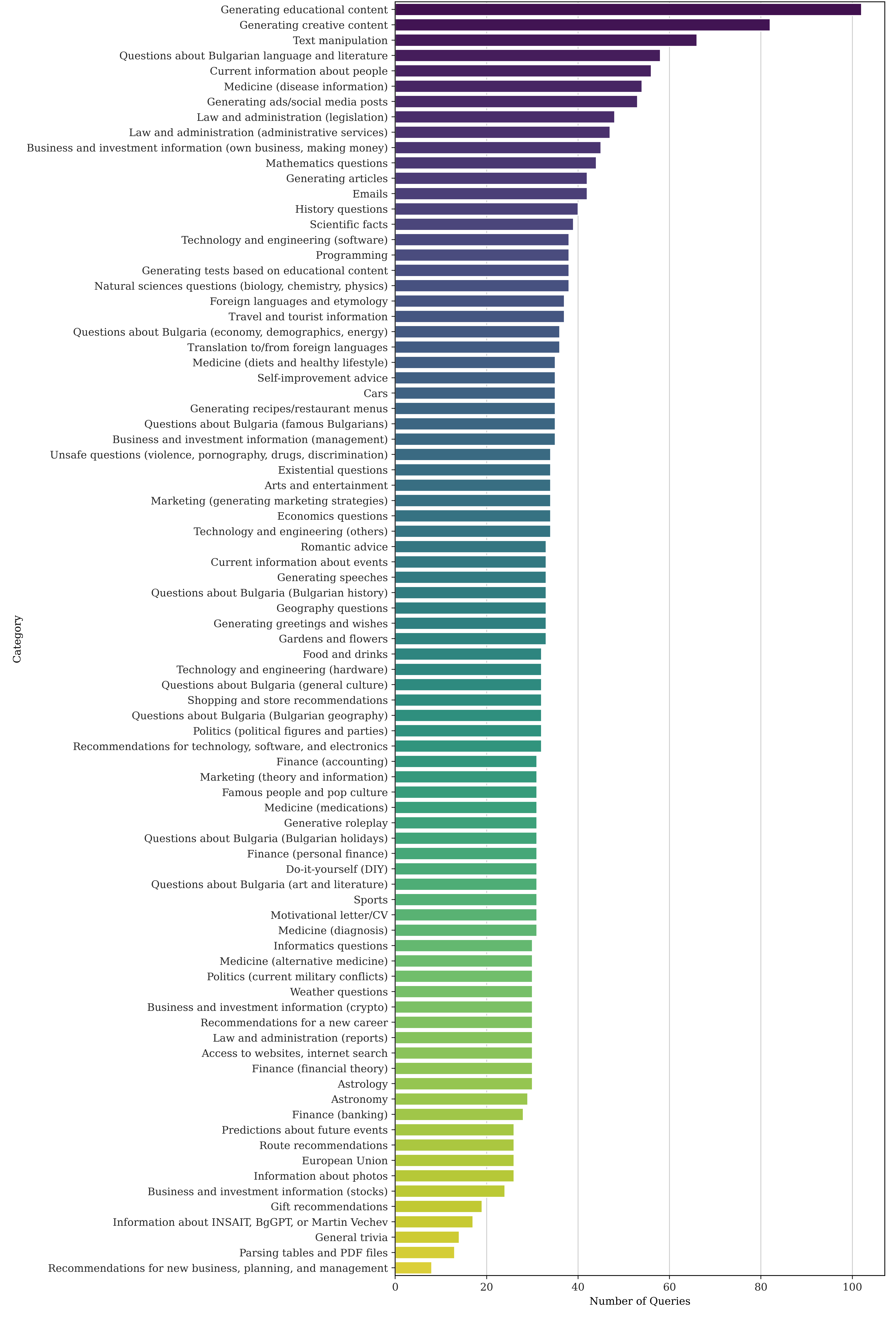} % Replace with your file
        \caption{Distribution of the 82 topics assigned as a main topic of the preference benchmark questions. The category "Other" is removed from the visualization.}
        \label{fig:chat_topics_ift}
    \end{minipage}
    \label{fig:chat_topics}
\end{figure*}

\begin{figure}[htbp]
    \centering
    \begin{subfigure}[t]{0.9\textwidth}
        \centering
        \includegraphics[width=\textwidth]{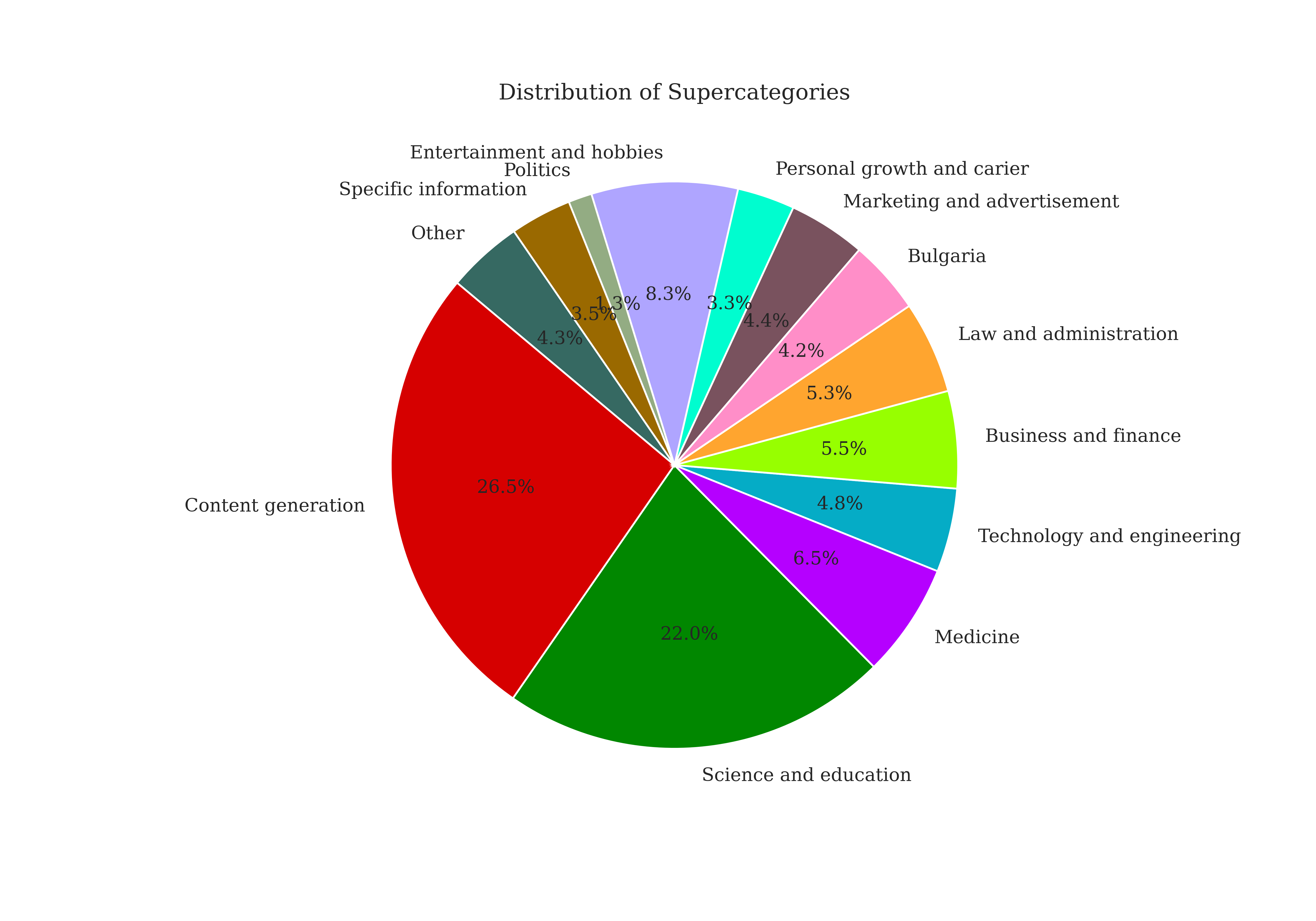}  % Replace with your file
        \caption{Distribution of the supercategories of the assigned-as-relevant topics to the queries from the chat IFT dataset.}
        \label{fig:chat_supecat_eval}
    \end{subfigure}
    \hfill
    \begin{subfigure}[t]{0.9\textwidth}
        \centering
        \includegraphics[width=\textwidth]{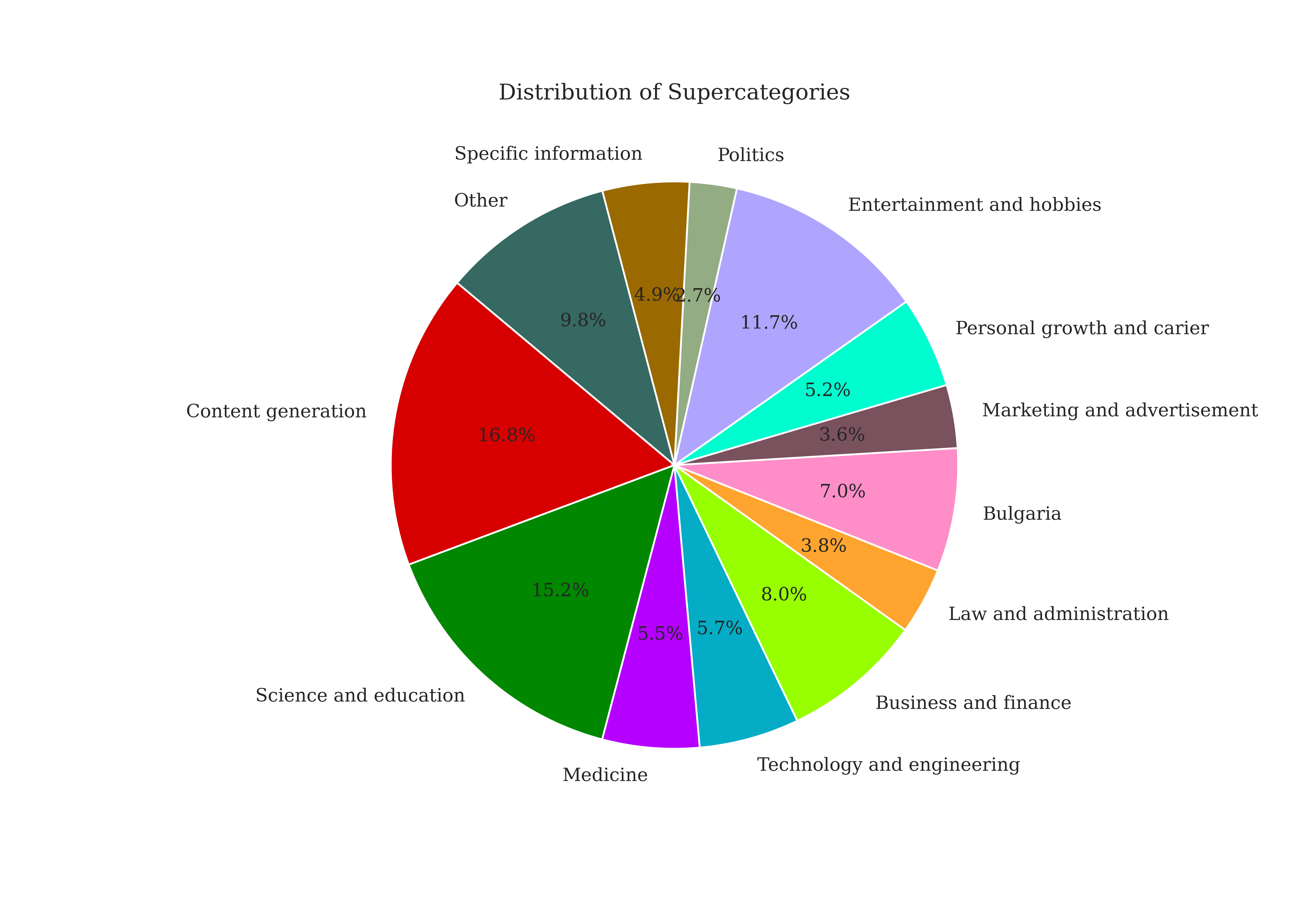} % Replace with your file
        \caption{Distribution of the supercategories of the assigned main topics of the preference benchmark questions.}
        \label{fig:chat_supercat_ift}
    \end{subfigure}
    \label{fig:chat_supercat}
\end{figure}

\end{document}